\documentclass[10pt,twocolumn,letterpaper]{article}

\usepackage{cvpr}              % To produce the CAMERA-READY version
\usepackage[accsupp]{axessibility}  % Improves PDF readability for those with disabilities.

\usepackage{times}
\usepackage{epsfig}
\usepackage{graphicx}
\usepackage{amsmath}
\usepackage{amssymb}
\usepackage{multirow}
\usepackage[dvipsnames]{xcolor}
\usepackage{booktabs}
\usepackage{array}
\usepackage{subcaption}
\usepackage[pagebackref,breaklinks,colorlinks]{hyperref}

\usepackage[capitalize]{cleveref}
\crefname{section}{Sec.}{Secs.}
\Crefname{section}{Section}{Sections}
\Crefname{table}{Table}{Tables}
\crefname{table}{Tab.}{Tabs.}

\def\cvprPaperID{3411} % *** Enter the CVPR Paper ID here
\def\confName{CVPR}
\def\confYear{2022}

\newcolumntype{P}[1]{>{\centering\arraybackslash}p{#1}}

\newcommand{\calL}{\mathcal{L}}

\newcommand{\calT}{\mathcal{T}}
\newcommand{\pose}{\text{hand-pose}}
\newcommand{\ThD}{\text{3D}}
\newcommand{\TwD}{\text{2D}}

\newcommand{\heatmap}{\mathcal{H}}

\newcommand{\featVec}{\mathcal{F}}

\newcommand{\twohodataset}{\text{H\textsubscript{2}O-3D}}

\begin{document}

%%%%%%%%% TITLE - PLEASE UPDATE
% \title{HandsFormer: Keypoint Transformer for Monocular 3D Pose Estimation of Hands and Object in Interaction}
% \title{\shreyas{Keypoint Transformer: Leveraging Keypoints for Joint-Specific Feature Selection and 3D Pose Estimation of Hands and Object in Interaction}}
\title{Keypoint Transformer: Solving Joint Identification in \\Challenging Hands and Object Interactions for Accurate 3D Pose Estimation}

\author{Shreyas Hampali\textsuperscript{(1)}, 
Sayan Deb Sarkar\textsuperscript{(1)}, Mahdi Rad\textsuperscript{(1)}, 
Vincent Lepetit\textsuperscript{(2,1)} \and
\textsuperscript{(1)}{\normalsize Institute for Computer Graphics and Vision, Graz University of Technology, Graz, Austria }\\ 
\textsuperscript{(2)}{\normalsize 
LIGM, Ecole des Ponts, Univ Gustave Eiffel, CNRS, Marne-la-Vall\'ee, France
} \\
{\tt\small \{<firstname>.<lastname>\}@icg.tugraz.at, vincent.lepetit@enpc.fr} \\
{\tt\small Project Website: \href{https://www.tugraz.at/index.php?id=57823}{https://www.tugraz.at/index.php?id=57823}}
% {\small Project page: \href{https://www.tugraz.at/index.php?id=50484}{ \color{blue} https://www.tugraz.at/index.php?id=50484}}
}
\maketitle

%%%%%%%%% ABSTRACT
\begin{abstract}

   We propose a robust and accurate method for estimating the 3D poses of two hands in close interaction from a single color image. This is a very challenging problem, as large occlusions and many confusions between the joints may happen. 
   State-of-the-art methods solve this problem by regressing a heatmap for each joint, which requires solving two problems simultaneously: localizing the joints \emph{and} recognizing them. In this work, we propose to separate these tasks by relying on a CNN to first localize joints as 2D keypoints, and on self-attention between the CNN features at these keypoints to associate them with the corresponding hand joint. The resulting architecture, which we call ``Keypoint Transformer'', is highly efficient as it achieves state-of-the-art performance with roughly half the number of model parameters on the InterHand2.6M dataset. We also show it can be easily extended to estimate the 3D pose of an object manipulated by one or two hands with high performance. Moreover, we created a new dataset of more than 75,000 images of two hands manipulating an object fully annotated in 3D and will make it publicly available.  
\end{abstract}

%%%%%%%%% BODY TEXT

\section{Introduction}

% \vincentrmk{"Keypoint Transformer" looks maybe better than keypoint-transformer?  People tend to use a capital T for Transformer}

\begin{figure}
\begin{center}
    \begin{tabular}{cc}
        \includegraphics[width=0.45\linewidth]{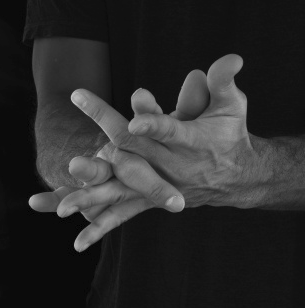} & \includegraphics[width=0.45\linewidth]{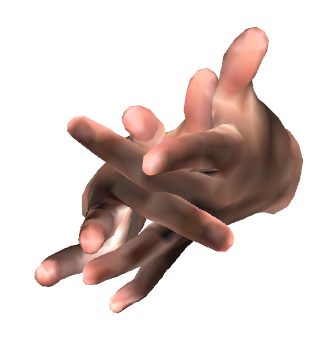} \\[0.3cm]
        \includegraphics[width=0.45\linewidth]{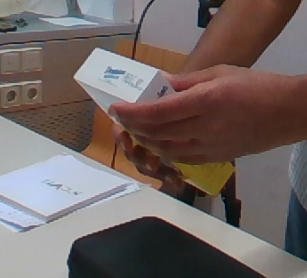} & 
        \includegraphics[width=0.45\linewidth]{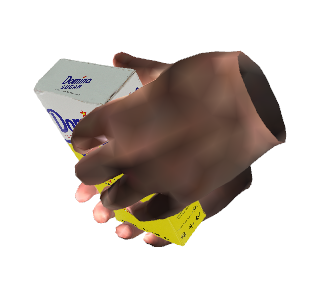} \\
    \end{tabular}
\end{center}
\vspace{-0.5cm}
\caption{Our approach accurately predicts 3D hand and object poses from a single RGB image in challenging scenarios including complex hand interactions (top) and 2 hands interacting with an object where the hands can be severely occluded (bottom). The bottom example is from the $\twohodataset$ dataset we  also introduce in this paper, which contains challenging, fully and accurately 3D-annotated,  video sequences of two hands manipulating objects.}
\label{fig:teaser}
\vspace{-0.5cm}
\end{figure}

3D hand pose estimation has the potential to make virtual reality, augmented reality, and interaction with computers and robots much more intuitive. Recently, significant progress has been made for single-hand pose estimation using depth maps and even single RGB images. Being able to deal with RGB images is particularly attractive as it does not require a power-hungry active sensor. Many approaches have been proposed, mostly based on direct prediction with different convolutional network architectures~\cite{umar,Zimmermann2017LearningTE,hasson19_obman,Moon2020I2LMeshNetIP,Adrian,Tekin2019HOUE,Panteleris2018UsingAS} of the 3D joint locations or angles, or relying on rendering for fine pose estimation and tracking~\cite{Baek2019PushingTE, Oberweger2015TrainingAF, hampali2020honnotate,Pavlakos2018LearningTE,Tung2017SelfsupervisedLO}.

\begin{figure}
\begin{center}
{\footnotesize
    \begin{tabular}{c@{\;}c@{\;}c@{\;}c}
        \includegraphics[height=0.225\linewidth]{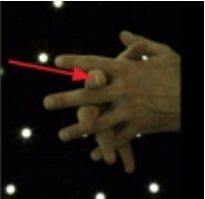} & 
        \includegraphics[height=0.225\linewidth]{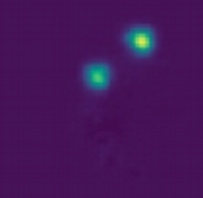} & 
        \includegraphics[height=0.225\linewidth]{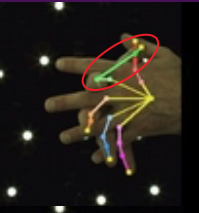} & 
        \includegraphics[height=0.225\linewidth]{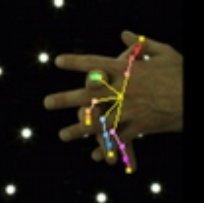} \\
        Input & Heatmap from \cite{Moon_2020_ECCV_InterHand2.6M}  & Right hand pose & Right hand pose \\
        image & for the & recovered & recovered \\
        &Right Index Tip & by \cite{Moon_2020_ECCV_InterHand2.6M} & by our method \\
    \end{tabular}
}    
\end{center}
\vspace{-0.5cm}
\caption{\textbf{Similar appearances between joints and partial occlusions make previous methods prone to failure. }
The InterNet state-of-the-art method \cite{Moon_2020_ECCV_InterHand2.6M} predicts a heatmap for each joint but the predicted heatmaps can become ambiguous, resulting in failures when predicting the hand pose (the hand on the back in this example). Our approach explicitly models the relationship between keypoints resulting in more accurate poses. More examples can be found in the supplementary material. }
\label{fig:ambiguities}
\label{fig:joint_confusion}
\vspace{-0.5cm}
\end{figure}

In contrast to single-hand pose estimation, two-hand pose estimation has received much less attention. This problem is indeed significantly harder: The appearance similarities between the joints of the two hands make their identification extremely challenging. Moreover, in close interaction, some of the joints of a hand are likely to be occluded by the other hand or itself. Thus, first detecting the left and right hands and then independently predicting their 3D poses~\cite{Han2020MEgATrackME, Panteleris2018UsingAS} performs poorly in close interaction scenarios.  Bottom-up approaches~\cite{Moon_2020_ECCV_InterHand2.6M, Wang2020RGB2HandsRT} directly estimate the 2D joint locations and their depths using one heatmap per joint. However, as shown in Fig.~\ref{fig:joint_confusion}, the similarity in appearances of the joints and severe occlusions degrade the quality of heatmaps failing to localize the joints accurately. More recent works~\cite{Dong,Baowen,fan2021digit} have attempted to alleviate this problem by exploiting joint-segmentation, joint-visibility, or by adding more refinement layers increasing the overall complexity of the network. By exploiting only keypoints, our method outperforms these methods by a large margin with a significantly smaller model.

% \shreyas{
% As shown in Fig.~\ref{fig:block_dig}, in this work, we estimate the 3D pose of the hands in three stages: 1) Keypoint Sampling: Keypoints are potential joint locations in the image and their locations are obtained from a single-channel heatmap. 2) Keypoint-Joint Association: The estimated keypoints are associated with different joints from the both the hands or to the background class in case of false positives. 3) Pose Estimation: For each joint, the corresponding keypoint features are selected and the 3D pose is estimated. 
% Our architecture termed as ``Keypoint Transformer'' is designed to disambiguate the identity of keypoints and allows for 3D pose estimation using different pose representations. Fig.~\ref{fig:teaser} shows the output when using MANO~\cite{Romero2017EmbodiedH} mesh as the output representation and Fig.~\ref{fig:joint_confusion} shows the output when using 2.5D~\cite{umar} pose representation.
% }

As shown in Fig.~\ref{fig:block_dig}, instead of aiming to localize and recognize the hand joints simultaneously, we estimate the 3D poses of the hands in three stages: 
(1) We first detect ``keypoints'', which are potential joint locations in the image, by predicting a \emph{single} heatmap. These keypoints do not have to exactly match all the hand joints: The 3D poses we predict are still correct if some joints are not detected as keypoints, and if some keypoints do not correspond to joints.
(2) Then, we associate the keypoints with the corresponding joint or to the background in the case of false positives, on the basis of the keypoint locations and their image features. This is done for all the keypoints simultaneously to exploit mutual constraints, using the self-attention mechanism. 
(3) Finally, we predict the 3D hand poses using a cross-attention module, which selects keypoints associated with each of the hand joints. Our approach is agnostic to the parameterization of the pose and we consider three different hand pose representations. 

Our architecture, which we call ``Keypoint Transformer'', is therefore designed to explicitly disambiguate the identity of the keypoints and performs very well even on complex configurations. Fig.~\ref{fig:teaser} shows its output on two challenging examples, using the MANO~\cite{Romero2017EmbodiedH} mesh as the output representation.
Our architecture is related to the ``Detection Transformer''~(DETR)~\cite{detr} architecture. DETR uses all the spatial features from a low-resolution CNN feature map, combined with learned location queries to detect objects in an image. The high computational complexity of the Transformer restricts DETR from using higher resolution CNN feature maps. As we show in our experiments, using the DETR-style architecture for hand pose estimation results in lower accuracy and we hypothesize that this is due to the use of lower resolution feature maps and features from the entire image.

% Several works have combined CNN and Transformer~\cite{Vaswani} architectures in the past for various Computer Vision problems~\cite{Khan2021TransformersIV,detr,zhu2021deformable,Dosovitskiy2020AnII,metro,Huang2020HandTransformerNS,Yang2020LearningTT,Kumar2021ColorizationT}. \cite{metro} regresses the mesh vertices of a human body or a hand from a single RGB image using multiple layers of self-attention. \cite{Huang2020HandTransformerNS} estimates the hand pose from point cloud data using encoder-decoder Transformer architecture. While these works are aimed at single hand pose estimation and their extension to two hands is non-trivial, our architecture is designed to estimate single and two-hands poses along with the object pose during hand-object interaction from the input RGB image. DETR~\cite{detr} is a CNN-Transformer architecture for object detection that is closely related to our architecture. DETR uses all the spatial features from a low-resolution CNN feature map, combined with learned location queries to detect objects in an image. The high computational complexity of the Transformer restricts DETR from using higher resolution CNN feature maps. As we show in our experiments, using the DETR-style architecture for hand pose estimation results in lower accuracy and we hypothesize that this is due to the use of lower resolution feature maps and features from the entire image.

We train and evaluate our architecture on the recent InterHand2.6M hand-hand~\cite{Moon_2020_ECCV_InterHand2.6M} and HO-3D hand-object~\cite{hampali2020honnotate} interaction datasets. We also introduce a challenging dataset of videos with two hands interacting with an object with complete and accurate 3D annotations without markers. This dataset is based on the work of \cite{hampali2020honnotate}, and we call it $\twohodataset$. Our method achieves state-of-the-art performance on existing hand-interaction datasets and serves as a strong baseline for the $\twohodataset$ dataset. Our experiments show that on InterHand2.6M, our method achieves state-of-the-art performance with roughly half the number of model parameters. We carry out several ablation studies and compare with strong baselines to prove the efficacy of our approach.

% \vincentrmk{We can remove this if needed:}

% The rest of the paper is organized as follows: We discuss related work in Section~\ref{sec:related_works} and detail our method in Section~\ref{sec:method}. In Section~\ref{sec:eval}, we provide the results of our evaluation on three different datasets, and perform a detailed ablation study in Section~\ref{sec:discussions}. We  conclude in Section~\ref{sec:conclusion}.

\section{Related Work}
\label{sec:related_works}

Many approaches have already been proposed for hand or object pose estimation from either RGB images or depth maps. Here we focus mainly on works that estimate hand poses during hand-hand or hand-object interactions. We also discuss recent advances in Transformer architectures for computer vision as they are highly relevant to our work.

%% Hand pose estimation is a widely researched topic with several approaches presented in the past for estimating 3D pose from RGB and depth images. Here we focus mainly on works that propose estimating hand poses during interactions i.e., hand-hand or hand-object interactions. We also discuss recent advances in transformer architectures in computer vision problems due to their strong relevance to our work.

\subsection{Interacting Hand Pose Estimation}

Hand pose estimation methods can be broadly classified as generative, discriminative, or hybrid approaches. Generative methods~\cite{Oikonomidis2012TrackingTA, Oikonomidis2011Tracking, Panteleris2015, Kyriazis2014Scalable3T, Tzionas20153DOR, hampali2020honnotate}  fit a parametric hand model to an observed image or depth map by minimizing a fitting error under some constraints. Discriminative methods~\cite{Panteleris2018UsingAS, Tekin2019HOUE, Zimmermann2019FreiHANDAD, hasson19_obman, hasson20_handobjectconsist, Karunratanakul2020GraspingFL, Brahmbhatt2020ContactPoseAD,Moon_2020_ECCV_InterHand2.6M} mostly directly predict the hand pose from a single frame. Generative methods often rely heavily on tracking and are prone to drift whereas discriminative methods tend to generalize poorly to unseen images~\cite{Armagan2020MeasuringGT}. Hybrid approaches~\cite{Ballan2012MotionCO, Taylor2016EfficientAP, Tzionas2016CapturingHI, Sridhar2016RealTimeJT, Taylor2017ArticulatedDF, Mueller2019RealtimePA, Xiang2019MonocularTC, Cao2020ReconstructingHI,Han2020MEgATrackME, Smith2020ConstrainingDH, Wang2020RGB2HandsRT} try to combine the best of these two worlds by using discriminative methods to detect visual cues in the image followed by model fitting.

Earlier methods~\cite{Oikonomidis2011Tracking, Kyriazis2014Scalable3T, Oikonomidis2012TrackingTA} for generative hand pose estimation during interaction used complex optimization methods to fit a parametric hand model to RGBD data. %\cite{Panteleris2015, Tzionas20153DOR} reconstruct the object model during in-hand scanning while continuously tracking the hand in a RGBD camera setup. 
\cite{hampali2020honnotate} proposed multi-frame optimization to fit hand and object models to RGBD data from multiple RGBD cameras. Generative methods alone often lose tracking during close interactions or occlusions and are hence combined with discriminative methods to guide the optimization.

\cite{Ballan2012MotionCO, Tzionas2016CapturingHI} detect the fingertips as discriminative points and used them in the optimization along with a collision term and physical modelling. Recently, \cite{Smith2020ConstrainingDH} proposed high-fidelity hand surface tracking of hand-hand interactions in a multi-view setup where the regressed 3D hand joint locations were used for initializing the tracking. \cite{Wang2020RGB2HandsRT, Mueller2019RealtimePA,Han2020MEgATrackME,Panteleris2018UsingAS, Cao2020ReconstructingHI} compute dense features or keypoints from a single RGB or depth image and fit a hand model~\cite{Romero2017EmbodiedH} to these estimates with physical constraints and joint angle constraints. Fully discriminative methods~\cite{Tekin2019HOUE,hasson19_obman,Moon_2020_ECCV_InterHand2.6M,hasson20_handobjectconsist} jointly estimate the 3D joint locations or hand model parameters of both the interacting hands or the interacting hand and the object by incorporating contacts and inter-penetrations in the training. \cite{Karunratanakul2020GraspingFL} estimates the hand-object surface using implicit representation that naturally allows modelling of the contact regions between hand and object. \cite{Dong,fan2021digit} improve the accuracy of 3D pose estimation during hand-hand interaction scenarios by incorporating joint-visibility and part-segmentation cues, whereas \cite{Baowen} utilize refinement layers to iteratively refine the estimated poses.

By contrast with the above mentioned approaches designed specifically for hand-hand or hand-object interaction scenarios, we propose in this work a unified discriminative approach for all hand interaction scenarios. Further, many previous discriminative methods perform poorly during close hand interactions due to similarity in appearance of the joints. In this work, we model relationship between all detected joints in the image resulting in more accurate pose estimation while keeping the model complexity low.

The success of discriminative methods depend on the variability of training data and several hand interaction datasets have been proposed.  \cite{GarciaHernando2018FirstPersonHA} first provided a marker-based hand-object interaction dataset using RGBD cameras.  %\cite{Zimmermann2019FreiHANDAD} proposed a RGB dataset with many hand-object interaction images but annotated only with the 3D poses for the hand.
\cite{hampali2020honnotate,Zimmermann2019FreiHANDAD} and \cite{hasson19_obman} respectively proposed real and synthetic hand-object interaction dataset with a single hand manipulating an object, while \cite{Kwon_2021_ICCV} recently developed a two-hands and object interaction dataset. \cite{Brahmbhatt2020ContactPoseAD} proposed single and two-hand object interaction dataset using infrared camera for contact annotations.
\cite{Moon_2020_ECCV_InterHand2.6M} developed a large-scale two-hand interaction dataset using semi-automatic annotation process with many close interactions. \cite{Taheri2020GRABAD} used MoCap to obtain pose of full body, hand, and object during interaction and used it to generate realistic grasp on unseen objects. %\cite{Brahmbhatt2020ContactPoseAD} used infrared cameras to identify single and two-hand contact regions on object and developed a markerless static hand-object interaction dataset with accurate contact annotations. More recently, %\cite{Kwon_2021_ICCV} developed a two-hands and object interaction dataset in a markerless setting using an automatic annotation method.

In this work, we also introduce a challenging two-hands-and-object interaction dataset which we created using the optimization method of \cite{hampali2020honnotate}.  Our dataset is made of videos of two hands from different subjects  manipulating an object from the YCB dataset~\cite{ycb}, annotated with the 3D poses of the hands and the object. Our architecture already performs well on this dataset and constitutes a strong baseline.

\subsection{Transformers in Computer Vision}
% \shreyasrmk{add graphformer}
%% \shreyas{Transformers are often used along with CNNs in vision related tasks \cite{} and has shown promising results in tasks such as detection \cite{}, classification \cite{}, image super-resolution \cite{} and colorization \cite{}.}

% Transformers are extremely effective in solving language-related tasks for which they were introduced~\cite{Vaswani} and have recently been increasingly gaining popularity for vision related problems~\cite{Khan2021TransformersIV}.  Attention mechanism~\cite{ParikhT0U16,Vaswani} which is at the core of the transformer architecture, takes as input a set of features of arbitrary length and transforms them into a output sequence of same length.

% In vision related problems, the features are often extracted from a CNN backbone and different architectures have been proposed to solve object detection~\cite{detr,zhu2021deformable}, image classification~\cite{Dosovitskiy2020AnII}, pose estimation~\cite{metro,Huang2020HandTransformerNS} and low-level image tasks~\cite{Yang2020LearningTT,Kumar2021ColorizationT}.  We refer the reader to \cite{Khan2021TransformersIV} for a detailed survey.

Transformers have recently been increasingly gaining popularity for vision related problems~\cite{Khan2021TransformersIV}.  Features are often extracted from a CNN backbone and different architectures have been proposed to solve object detection~\cite{detr,zhu2021deformable}, image classification~\cite{Dosovitskiy2020AnII}, pose estimation~\cite{metro,Huang2020HandTransformerNS,lin2021,guillem2021} and low-level image tasks~\cite{Yang2020LearningTT,Kumar2021ColorizationT}.  We refer the reader to \cite{Khan2021TransformersIV} for a detailed survey.

% \cite{detr} proposed to combine CNN backbone with transformer and showed that the resulting model achieves competitive performance when compared to well-established object detection methods while avoiding the need for hand-designed components. \cite{zhu2021deformable} further improved \cite{detr} by using multi-scale features from the CNN backbone and restricted the attention to only a small set of features from sampled locations. The sampling locations were regressed on-the-fly from the query features of the attention module. 
% \cite{metro} proposed to reconstruct the human body or hand mesh vertices from a single RGB image using multiple transformer encoder layers and achieved state-of-the-art performance. However, \cite{metro} reconstructs an isolated single body or hand from an image. \cite{Huang2020HandTransformerNS} estimate 3D pose from hand point-cloud data using transformer encoder-decoder architecture and proposed to generate query embeddings from input point-cloud instead of learning them as in \cite{detr, zhu2021deformable}. Our proposed architecture has 2 key differences from previous works 1) we provide multi-scale image features sampled from keypoint locations as input tokens to the transformer 2) our learned query embeddings represent joint identities and not positional features as in previous works. We show in our experiments that this novel architecture enables more accurate pose estimation. 

\cite{detr, zhu2021deformable} proposed to combine a CNN backbone with a Transformer to detect objects in an image. 
\cite{metro} proposed to reconstruct the vertices of a single human body or hand from an RGB image using multiple Transformer encoder layers and achieved state-of-the-art performance. \cite{lin2021} improved \cite{metro} by using graph convolutions along with a Transformer encoder. \cite{Huang2020HandTransformerNS} estimated a 3D pose from hand point-cloud data using a Transformer encoder-decoder architecture and proposed to generate query embeddings from input point-cloud instead of learning them as in \cite{detr, zhu2021deformable}. While these works are aimed at single hand pose estimation and their extension to two hands is non-trivial, our architecture is designed to estimate single and two-hands poses along with the object pose during hand-object interaction from the input RGB image.

In a closely related work, \cite{guillem2021} solves the multi-person 2D pose estimation problem using detected keypoints and person centers, by associating the joint keypoints to the correct person center using attention. There are however several key differences: In our case, the hand centers get very close to each other during close interaction, and the approach in \cite{guillem2021} would not be transferable. More importantly,  hand joints are much more ambiguous than ``body joints'' as they look very similar to each other. Our method is also robust to undetected and falsely detected keypoints as we show in our discussions, while \cite{guillem2021} cannot handle undetected keypoints. Further, we show that, by randomly sampling keypoints on the object, we can easily extend our method to 3D object pose estimation during hand-object interactions.

% \shreyas{In a concurrent work, \cite{guillem2021} solved the multi-person 2D pose estimation problem using detected keypoints and person centers. The joint keypoints and center keypoint belonging to each person in the image were grouped using attention scores and the 2D pose was estimated. A similar approach cannot be extended to interacting hands as the hand centers are very close to each other during close interaction. Further, our approach estimates the pose in 3D and also extends to object pose estimation during hand-object interactions.}

% \shreyas{\cite{guillem2021} solved the multi-person 2D pose estimation problem by using Transformer encoder to group the detected 2D joints for each person. Different from these previous methods, our architecture uses only a set of keypoints from a single-channel heatmap and associate them with different joints of both the hands, separating the keypoint localization and keypoint identification tasks. The identity-aware keypoint features are used for 3D pose estimation.} \vincentrmk{the difference is not clear}
%Different from these previous methods, our Keypoint Transformer architecture enables accurate 3D hands and object pose estimation by selecting localized CNN features guided by the keypoints.}

%Different from these previous architectures, our method samples multi-scale image features from keypoint locations and uses a Transformer encoder-decoder architecture to estimate 3D pose of interacting hands and object from the sampled features.

\section{Method}
\label{sec:method}

\begin{figure*}[t!]
\begin{center}
  \includegraphics[trim=60 125 60 64,clip, width=0.8\linewidth]{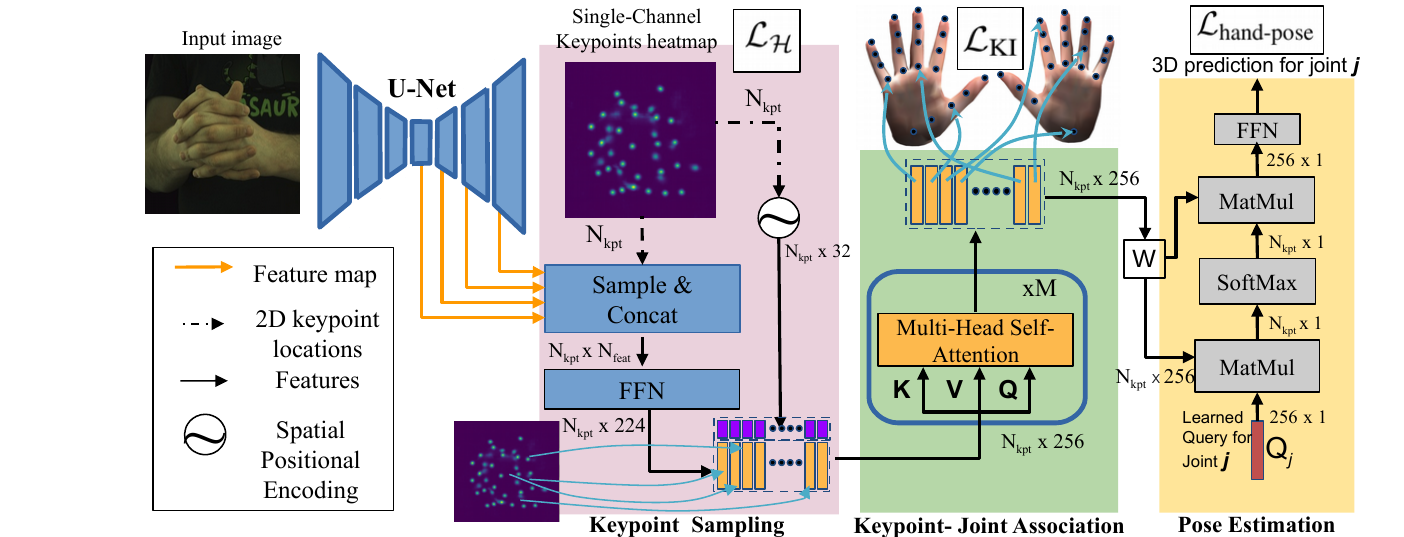} 
\end{center}
\vspace{-0.5cm}
\caption{\textbf{Overview of our approach. } We detect keypoints which are potential locations of joints and encode them with CNN image features and spatial embedding~(Section~\ref{sec:keypoints}). From this information, the self-attention module creates context-aware keypoint features which are essential for associating each keypoint with the corresponding joint~(Section~\ref{sec:assoc}). A cross-attention module finally predicts for each joint (using learned queries) the values required for computing the hand poses~(Section~\ref{sec:pose_est}). The exact nature of these values depends on the chosen representation of the hand pose~(Section~\ref{sec:pose_rep_loss}). Not all the keypoints have to correspond to a joint and not all the joints have to be detected as keypoints, which makes our approach very robust but still accurate as it relies on keypoints (as discussed in Section~\ref{sec:related_works}).}
\vspace{-0.5cm}
\label{fig:block_dig}
\end{figure*}

As shown in Fig.~\ref{fig:block_dig}, our architecture first detects keypoints that are likely to correspond to the 2D locations of hand joints and encodes them as input to the keypoint-joint association stage. The keypoints are encoded with their spatial locations and the image features at these locations. The self-attention layers in the Keypoint Transformer disambiguate the keypoints and associates them with different joint types and a background class. The (single) cross-attention layer then selects these ``identity-aware keypoints'' to predict root-joint-relative pose parameters of both hands, plus additional parameters such as the translation between the hands and hand shape parameters.

We detail below the keypoint detection and encoding step, how we use the Keypoint  Transformer to predict the hands poses, the representations we considered for the 3D hand poses, and the loss used for training. We also explain how our approach can be extended to object pose estimation during hand-object interaction scenarios.

\subsection{Keypoint Detection and Encoding}
\label{sec:kpt_encode}
\label{sec:keypoints}
Given the input image, we first extract keypoints that are likely to correspond to 2D hand joint locations.  To do this, we predict a single-channel heatmap  $\heatmap$ from the input image using a standard U-Net~\cite{unet} architecture, and we keep its local maximums using a non-differentiable, non-maximum suppression operation. 
At this stage, we do not attempt to recognize which keypoint corresponds to which joint as it is a difficult task, and the predicted heatmap has only one channel.
In practice, we keep a maximum of $N_\text{hand}$ keypoints, with $N_\text{hand}=64$, while the number of hand joints is 42 in total for 2 hands. The 2D keypoint locations are normalized to $[0,1]$ range.

The ground truth heatmap $\heatmap^*$ is obtained by applying a 2D Gaussian kernel of variance $\sigma$ at each of the ground truth 2D joint locations and the U-Net is trained to predict the heatmap by minimizing the L2 loss.

We compute for each detected keypoint an appearance and spatial encoding to represent the keypoints as input to the next stage. As shown in Fig.~\ref{fig:block_dig}, for the appearance part, we extract image features from the decoder of the U-Net network. More exactly, we sample feature maps at multiple layers of the U-Net decoder at the normalized keypoint locations using bilinear interpolation and concatenate them to form a 3968-D feature vector, which is then reduced down to a 224-D encoding vector using a 3-layer MLP. For the spatial encoding, we obtain 32-D sine positional encoding similar to \cite{detr} corresponding to the 2D location of the keypoint. We finally concatenate the appearance and spatial encodings  to  form a 256-D vector representation of the keypoint. The keypoint detector is pre-trained before fine-tuning it jointly with the rest of the pipeline.

%%%%%%%%%%%%%%%%%%%%%%%%%%%%%%%%%%%%%%%%%%%%%%%%%%%%%%%%%%%%%%%%%%%%%%%%%%%%%%%%%%%%%%%%%%%%%%%

\subsection{Keypoint-Joint Association}
\label{sec:ki_loss}
\label{sec:assoc}

For each keypoint $K_i$, we have now an encoding vector $\featVec_i \in \mathbb{R}^{256}$. We use these vectors as input to the multi-layer, multi-head self-attention module with $N_{SA}$ layers. The self-attention~\cite{Vaswani} helps to model the relationship between the keypoints and create global context-aware feature $\mathcal{G}_i \in \mathbb{R}^{256}$, for each keypoint. Such context-aware features are necessary to associate the keypoints with different joint types using a ``keypoint-joint association'' loss we denote $\calL_\text{KI}$. As a result of $\calL_\text{KI}$, the keypoint features also now encode the joint identity information along with the localized CNN image features. 

The identity of keypoint $k$ is defined by $(h_k, j_k)$, where $h_k$ is the hand identity (left or right) and $j_k$ is the joint index. We also use an additional `background' identity for keypoints that are falsely detected. The keypoint identity is predicted using a feed-forward network~(FFN) consisting of a 2-layer MLP, a linear projection layer and a softmax layer. We use the cross-entropy loss for $\calL_\text{KI}$:
\begin{equation}
\calL_\text{KI} = \sum_i \text{CE}((h_i,j_i), (h_i^*,j_i^*)) \> ,
\end{equation}
where $(h_i^*,j_i^*)$ are the ground truth identities and $\text{CE}$ denotes the cross-entropy loss. To obtain the ground truth identity for the detected keypoints, we associate them at training time with the closest reprojection of a ground truth 3D joint, if the distance is below a threshold $\gamma$. If there are no reprojected joints within a distance of $\gamma$, the keypoint is assigned to the background class. We empirically set $\gamma = 3$~pixels in our experiments. Similar to  \cite{detr}, the keypoint identities are predicted after each layer of self-attention module using FFNs with shared weights and the loss is applied to predictions of each layer.

% The keypoint-joint association loss is a cross-entropy loss and is defined as:

The prediction can result in multiple keypoints assigned to the same joint identity and some keypoints assigned to the background class. As we discuss in Section~\ref{sec:discussions}, the keypoints associated to the background are ignored, while all the keypoints associated with a given joint are considered for estimating the pose of the corresponding joint by the cross-attention module.

\subsection{Pose Estimation from Identity-Aware Keypoints}
\label{sec:pose_est}
The keypoint-joint association loss enables the keypoint features to also encode joint identity information along with the image features and spatial embeddings. We use a single cross-attention layer with learned joint queries to predict which keypoint(s) match the queried joint identities. The cross-attention operation~\cite{Vaswani} for a learned joint query $Q_j \in \mathbb{R}^{256}$ and features $\{\mathcal{G}_i\}_i$ is given by
\begin{equation}
    \text{CA}\big(Q_j, {G}\big) = \text{softmax}\left(\frac{Q_j^TW_K{G}}{16}\right)(W_V G)^T \> ,
\end{equation}
where $G$ is a matrix whose columns contain feature vectors $\{\mathcal{G}_i\}_i$, and $W_K$ and $W_V$ are learnable matrices of dimension $256\times256$. Similar to \cite{Vaswani}, the cross-attention features are added to $Q_j$ to create a residual connection. The resulting features are transformed by a 3-layer MLP to map them to the pose space. 

The number of joint queries depend on the pose representation. We consider 3 different representations and describe them in Section~\ref{sec:pose_rep_loss} and the suppl. mat. For example, we use 21 joint queries for each of the 21 joints per hand when using 2.5D pose representation. Along with the joint queries, one for each joint of the two hands, we use an additional learned query to predict the relative translation $T_{L\rightarrow R}$ between the hands, and the 10-D MANO hand shape parameters $\beta$. These are learned using the L1 loss. The MANO shape parameters are useful when predicting the pose using the MANO joint angle representation.

\subsection{Hand Pose Representations and Losses}
\label{sec:pose_rep_loss}

We consider three main hand pose representations: 3D joint locations, 2.5D~\cite{umar, Moon_2020_ECCV_InterHand2.6M} joint locations, and MANO joint angles~\cite{Romero2017EmbodiedH}. Previous methods~\cite{hasson20_handobjectconsist, Angjoo, Pavlakos2018LearningTE, Pavlakos2019TexturePoseSH, hasson19_obman} have noted that regressing model parameters such as joint angles is less accurate in terms of joint error than regressing the joint locations directly. However, regressing MANO joint angles provides access to the complete hand mesh required for modeling contacts and interpenetration during interactions~\cite{Taheri2020GRABAD, Brahmbhatt2020ContactPoseAD, hasson19_obman} or for learning in a weakly supervised setup~\cite{Kulon2020WeaklySupervisedMH,Baek2020WeaklySupervisedDA,hasson20_handobjectconsist}, which could be interesting for future extension of our method. As we show later in our experiments, the Keypoint Transformer enables the MANO joint angle representation to achieve competitive performance when compared to the joint location representation. We follow standard practice for these three representations. We detail them and their corresponding losses in the supplementary material for completeness.

\begin{figure}
\begin{center}
\includegraphics[trim=40 100 220 100,clip, width=0.8\linewidth]{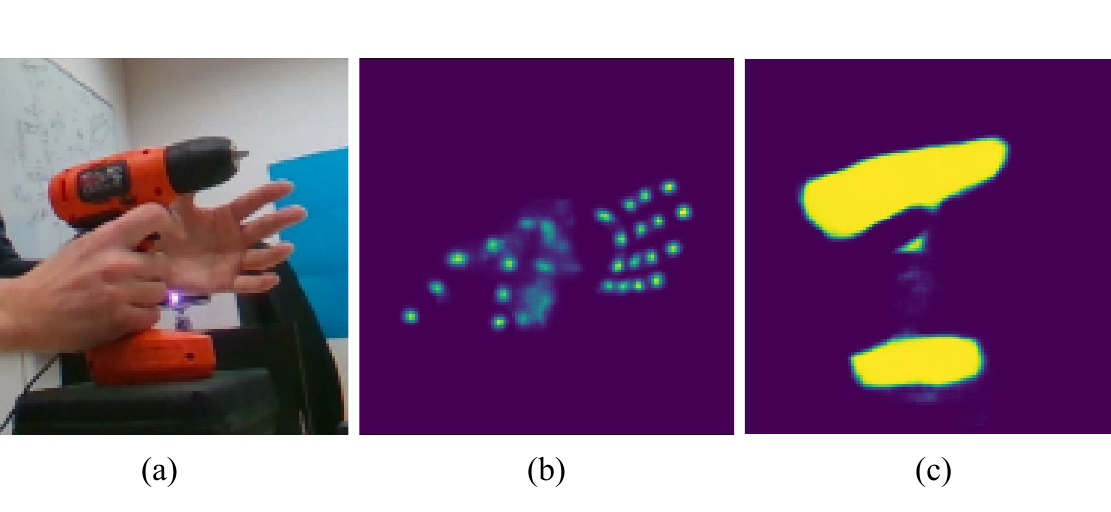} 
\end{center}
\vspace{-0.7cm}
\caption{\textbf{Keypoint detection for hands and object. } We train a U-net decoder to predict (b) a heatmap for all the joints together and (c) a segmentation map for the object from each we extract keypoints at random locations.}
\label{fig:heatmap}
\vspace{-0.5cm}
\end{figure}

\subsection{Object Pose Estimation}

Our method generalizes easily to predict the 3D pose of an object together with 3D poses of hands. As shown in Fig.~\ref{fig:heatmap}, along with the heatmap for the hand keypoints, we also predict a segmentation map of the object by adding an additional prediction head to the U-Net decoder. We then randomly select $N_\text{obj}=20$ points from this segmentation map and refer to them as `object keypoints'. We also tried estimating the heatmap of 2D reprojections of fixed points on the object mesh and selecting their local maximums as object keypoints and obtained similar results. 

We encode the appearance and spatial locations of the object keypoints in a 256-D vector, exactly like the hand keypoints. Collectively, these keypoint encodings cover the object appearance, allowing us to predict the 3D rotation and translation of the object. The encodings of $N_\text{obj}$ object keypoints and $N_\text{hand}$ hand keypoints are provided \emph{together} to the self-attention module.

Along with the hand keypoint identities $(h_k, j_k)$ and the background identity described in Section~\ref{sec:ki_loss}, we rely on an additional identity for the object. During the keypoint association stage, all the keypoints originating from the object are associated with the `object' identity, allowing the cross-attention module to only attend to object keypoints when estimating the object pose. Along with the joint queries that estimate the hand pose, we consider 2 additional queries in the cross-attention module and predict the 3D object rotation and 3D object translation relative to the right hand in a manner similar to that of hand pose. The object rotation is parameterized using the method proposed in \cite{Zhou2019OnTC}.

% and used in particular in \cite{cosypose}.

% Along with the joint queries that estimate the hand pose, we consider 2 additional queries in the cross-attention module for predicting the 3D object rotation and 3D object translation relative to the right hand. We parameterize the object rotation using the method proposed in \cite{Zhou2019OnTC} and used in particular in \cite{cosypose}. On the encoder side, similar to predicting the joint identities of the hand keypoints from their features, we also predict if a keypoint belongs to the object or not, allowing the transformer decoder to further differentiate between hand and object features during cross-attention.

We use a symmetry-aware object corner loss similar to \cite{Park2019Pix2PosePC} to train the network:
\begin{equation}
    \calL_\text{obj-pose} = \min_{R \in \mathcal{S}} \frac{1}{8} \sum_{i=1}^8 \lvert \lvert P\cdot B_i - P^* \cdot R \cdot B_i \lvert \lvert_2^2 \> ,
\end{equation}
\noindent where $P$ and $P^*$ denote the estimated and ground-truth object poses, $B_i$ denotes the $i^\text{th}$ corner of the 3D bounding box for the object in rest pose, and $\mathcal{S}$ is the set of rotation matrices which, when applied to the object, does not change its appearance.

%%%%%%%%%%%%%%%%%%%%%%%%%%%%%%%%%%%%%%%%%%%%%%%%%%%%%%%%%%%%%%%%%%%%%%%%%%%%%%%%%%%%%%%%%%%%%%%

\subsection{End-to-End Training}
We train our architecture end-end by minimizing the sum of the losses introduced above:
\begin{equation}
    \calL = \calL_\heatmap + \calL_\text{KI} + \calL_\calT + \calL_\pose +  \calL_\text{obj-pose}\>,
\end{equation}
where $\calL_\pose$ is the loss on the hand poses (detailed in the suppl. mat.) and $\calL_\calT$ is the L1 loss for relative translation between the two hands. Note that the keypoint detector is pretrained before training the entire network end-to-end. More optimization details are also given in the suppl. mat.

\section{Evaluation}
\label{sec:eval}

We evaluated our method on three challenging hand interaction datasets: InterHand2.6M, HO-3D, and our $\twohodataset$ dataset we introduce with this paper. We discuss them below.

\subsection{InterHand2.6M}
\label{sec:interhands}

\begin{figure*}[h!tb]
\begin{center}
\begin{minipage}{.125\linewidth}
\centering
\subfloat{\includegraphics[width=1\linewidth]{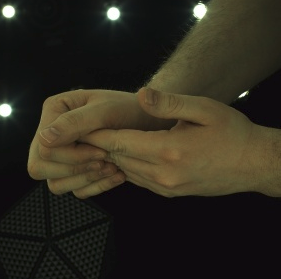}}
\end{minipage}%
\begin{minipage}{.125\linewidth}
\centering
\subfloat{\includegraphics[width=1\linewidth]{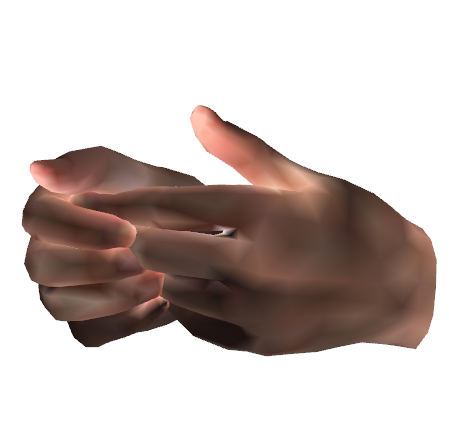}}
\end{minipage}%
\begin{minipage}{.125\linewidth}
\centering
\subfloat{\includegraphics[width=1\linewidth]{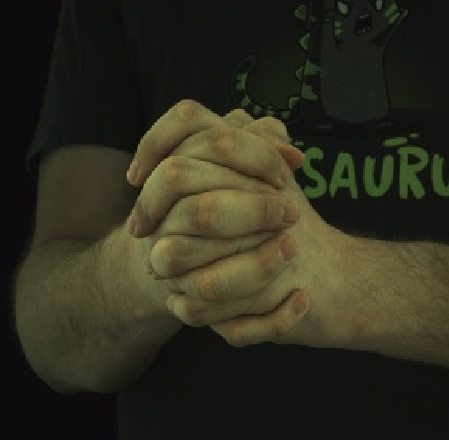}}
\end{minipage}%\par\medskip
\begin{minipage}{.125\linewidth}
\centering
\subfloat{\includegraphics[width=1\linewidth]{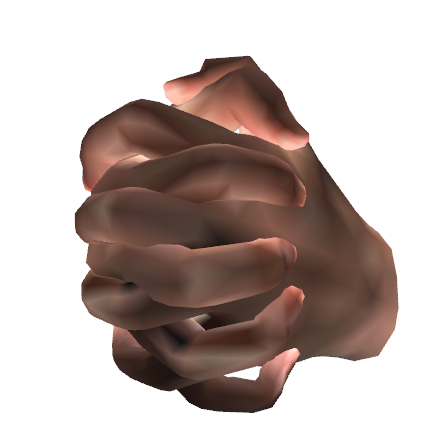}}
\end{minipage}%
\begin{minipage}{.125\linewidth}
    \centering
    \subfloat{\includegraphics[width=1\linewidth]{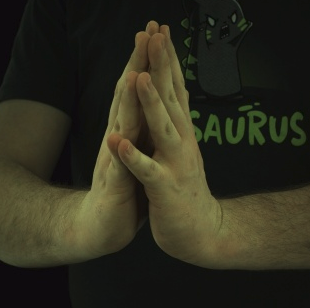}}
\end{minipage}%
\begin{minipage}{.125\linewidth}
    \centering
    \subfloat{\includegraphics[width=1\linewidth]{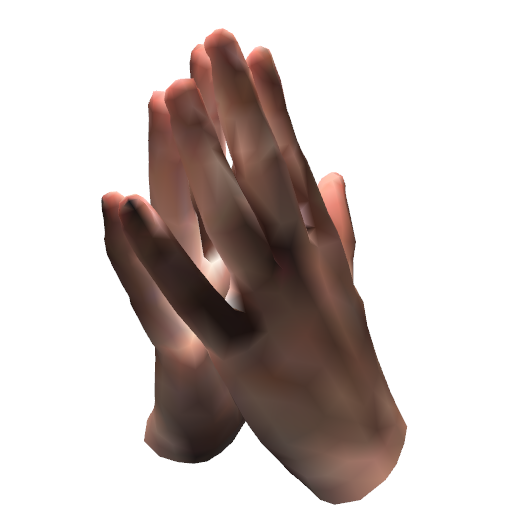}}
\end{minipage}%
% \begin{minipage}{.1\linewidth}
%     \centering
%     \subfloat{\includegraphics[width=1\linewidth]{figures/teaser/mano_img_4.png}}
% \end{minipage}%
% \begin{minipage}{.1\linewidth}
%     \centering
%     \subfloat{\includegraphics[width=1\linewidth]{figures/teaser/mano_pose_4.png}}
% \end{minipage}%
\begin{minipage}{.125\linewidth}
    \centering
    \subfloat{\includegraphics[width=1\linewidth]{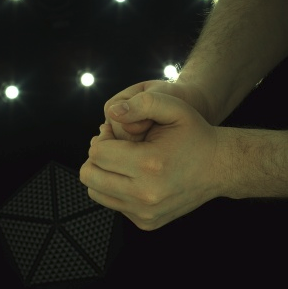}}
\end{minipage}%
\begin{minipage}{.125\linewidth}
    \centering
    \subfloat{\includegraphics[width=1\linewidth]{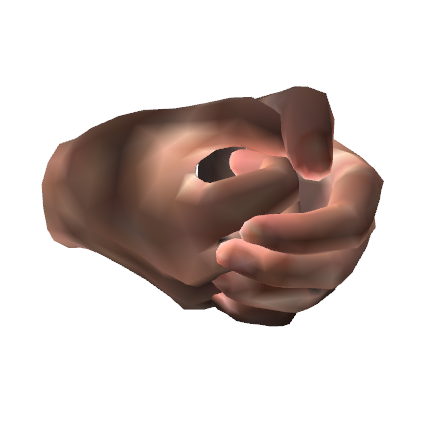}}
\end{minipage}%
% \begin{minipage}{.08\linewidth}
%     \centering
%     \subfloat{\includegraphics[width=1\linewidth]{figures/teaser/mano_img_4.png}}
% \end{minipage}%
% \begin{minipage}{.08\linewidth}
%     \centering
%     \subfloat{\includegraphics[width=1\linewidth]{figures/teaser/mano_pose_4.png}}
% \end{minipage}
\end{center}
\vspace{-0.5cm}
\caption{\textbf{Qualitative results on InterHand2.6M~\cite{Moon_2020_ECCV_InterHand2.6M}.} Our method obtains accurate poses of hands during complex interactions. We show the estimated MANO model from a different view.}
\label{fig:inter_results}
\vspace{-0.3cm}
\end{figure*}

\begin{table}[b]
    \centering
    \scalebox{0.72}{
    \begin{tabular}{@{}>{\raggedright}P{3.8cm}P{1.8cm}P{1.6cm}P{1.0cm}P{1.5cm}@{}}
    \toprule
    & \multicolumn{3}{c}{\centering MPJPE (mm)} & MRRPE\\
    & Single Hand & Two Hands & All & (mm)\\
    \midrule
    CNN+SA & 13.53 & 16.87 & 15.31 & 33.84 \\
    CNN+SA+CA (DETR~\cite{detr}) & 12.81 & 15.94 & 14.48 & 32.87 \\
    InterNet~\cite{Moon_2020_ECCV_InterHand2.6M} & 12.16 & 16.02 & 14.22 & 32.57 \\
    Ours & \textbf{10.99} & \textbf{14.34} & \textbf{12.78} & \textbf{29.63} \\
    \midrule
    Dong et al.~\cite{Dong} & - & - & 12.08 & - \\
    Ours & 9.10 & 11.98 & \textbf{11.30} & 21.89 \\
    \midrule
    Fan et al.~\cite{fan2021digit} & 11.32 & 15.57 & - & \textbf{30.51} \\
    Ours & \textbf{11.08} & \textbf{15.33} & 13.41 & 30.87 \\
    \bottomrule
    \end{tabular}
    }
    \vspace{-0.2cm}
    \caption{\textbf{Comparison with 2 baselines and the state-of-the-art methods on  InterHand2.6M~\cite{Moon_2020_ECCV_InterHand2.6M}.} We compare with \cite{Moon_2020_ECCV_InterHand2.6M,Dong,fan2021digit} using the different train/test splits reported in their works.}
    \label{tab:interhand}
    \vspace{-0.3cm}
\end{table}

\paragraph{Training and test sets.} InterHand2.6M~\cite{Moon_2020_ECCV_InterHand2.6M} is a recently published two-hand interaction dataset with many challenging poses. It was annotated semi-automatically and contains 1.36M train images and 849K test images.

\vspace{-12px}

\paragraph{Metrics.}
We report the Mean Per Joint Position Error~(MPJPE) and the Mean Relative-Root Position Error~(MRRPE) to evaluate the root-relative hand pose and the translation between the hands respectively, as in \cite{Moon_2020_ECCV_InterHand2.6M}.

\vspace{-12px}

\paragraph{Baselines.} We consider two Transformer-based baseline architectures. The first baseline~(`CNN+SA') provides the low-resolution~(32$\times$ downsampled) CNN feature maps after flattening along the spatial dimensions as input to the Transformer encoder containing self-attention~(SA) modules. The output tokens of the encoder are concatenated and the pose is predicted using an MLP. The second baseline~(`CNN+SA+CA') is more similar to DETR~\cite{detr}, where the low-resolution CNN feature maps are provided to the Transformer encoder-decoder architecture. The Transformer decoder contains SA and cross-attention~(CA) modules. The queries in the decoder are learnt and the pose is predicted using FFN similar to our Keypoint Transformer. We provide more details about the baselines in the suppl. mat. These baselines help to understand the importance of keypoint sampling and selection for pose estimation.

% As was done in \cite{Moon_2020_ECCV_InterHand2.6M} for evaluating their baseline, we consider the Mean Per Joint Position Error~(MPJPE) and the Mean Relative-Root Position Error~(MRRPE). MPJPE computes the Euclidean distance between the predicted and ground truth 3D joint locations after root joint alignment and indicates the accuracy of root-relative 3D pose. The alignment is carried out separately for the right and the left hands. MRRPE evaluates the accuracy of the localization of the left hand relative to the right hand. 

\vspace{-12px}

\paragraph{Results.} Table~\ref{tab:interhand} compares the accuracy of our method with the state-of-the-art method InterNet~\cite{Moon_2020_ECCV_InterHand2.6M}, and the two baselines, when using the 2.5D pose representation. Our method achieves 10\% higher accuracy than InterNet, which is a CNN-based architecture, and 16\% and 12\% higher accuracy than the two baselines, respectively. The higher accuracy of `CNN+SA+CA' w.r.t `CNN+SA' baseline demonstrates that soft-selection of image features by the decoder improves the accuracy. Further, the higher accuracy (12\%) of our Keypoint-Transformer w.r.t the `CNN+SA+CA' architecture shows that use of keypoint features for pose estimation instead of features from the entire image increases the overall accuracy.
% localized CNN image features selection guided by the keypoint detection increases the overall accuracy.

We compare our method with \cite{Dong} and \cite{fan2021digit} using their train and test splits. \cite{Dong,fan2021digit} use per-joint heatmaps coupled with joint visibility and segmentation guided features to improve the accuracy of the pose estimation, thus resulting in the model complexity that is higher than InterNet~\cite{Moon_2020_ECCV_InterHand2.6M}. Our method with a model complexity same as \cite{Moon_2020_ECCV_InterHand2.6M} (see Section~\ref{sec:discussions}) still outperforms these state-of-the-art methods. We show qualitative results in Fig.~\ref{fig:inter_results} and in the suppl. mat.

% \begin{table}[]
%     \centering
%     \scalebox{0.7}{
%     \begin{tabular}{P{2.0cm}|P{1.5cm}| P{1.8cm} | P{1.6cm} |  P{1.0cm} | P{1.0cm} }
%     & \multirow{2}{*}{\parbox{1.4cm}{\centering Camera Intrinsics}} & \multicolumn{3}{c | }{\centering MPJPE (mm)} &  \multirow{2}{*}{\parbox{1.0cm}{\centering MRRPE (mm)}}  \\
%     & & Single Hand & Inter. Hand & All &  \\
%     \hline
%     InterNet \cite{Moon_2020_ECCV_InterHand2.6M} & Yes & 13.79 & 21.24 & 17.54 & 40.46 \\
%     \hline
%     Joint Vec. & No & 12.42 & \textbf{17.08} & 14.76 & \textbf{33.14} \\
%     Joint Ang.* & No & 14.00 & 19.16 & 16.61 & 37.91 \\
%     Joint Ang. & No & 15.36 & 20.61 & 18.01 & 37.91 \\
%     2.5D Pose & Yes & \textbf{11.73} & 17.69 & \textbf{14.73} & 34.40 \\
%     \end{tabular}
%     }
%     \vspace{-0.2cm}
%     \caption{\textbf{Accuracy of our method with 3 different pose representations on  InterHandV0.0.} Our method achieves 16\% higher accuracy than \cite{Moon_2020_ECCV_InterHand2.6M} which relies on a fully CNN architecture. Even while estimating MANO~\cite{Romero2017EmbodiedH} joint angles our method outperforms \cite{Moon_2020_ECCV_InterHand2.6M} which estimates 3D joint locations directly. * indicates ground-truth 3D joints obtained from fitted MANO models.}
%     \label{tab:interhand}
%     \vspace{-0.3cm}
% \end{table}

\vspace{-0.1cm}

\subsection{HO-3D}
\label{sec:ho3d}

\begin{figure*}
\centering
\begin{tabular}{c@{}|c}
\includegraphics[trim=400 107 210 125, clip, height=0.35\linewidth]{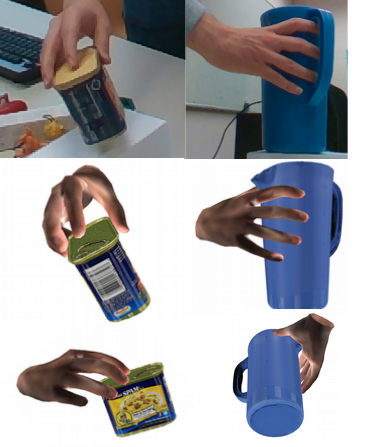}&
\includegraphics[trim=20  107 400 125, clip, height=0.35\linewidth]{figures/H2O3D_vis/h2o3d_ho3d_qual_small.pdf}
\\
HO-3D & $\twohodataset$\\
\end{tabular}
\vspace{-0.3cm}
\caption{\textbf{Qualitative results for our method on the $\twohodataset$ and HO-3D datasets}. Our method recovers poses even under large occlusions by the object and achieves state-of-the-art results on HO-3D  while serving as a strong baseline for our new dataset $\twohodataset$. Note that some objects (columns 2\&4) are considered to be rotationally symmetric along the z-axis.}
\vspace{-0.5cm}
\label{fig:ho_qual}
\end{figure*}

\paragraph{Training and test sets. } The HO-3D~\cite{hampali2020honnotate} dataset contains automatically annotated hand-object interaction sequences of a right hand and an object from the YCB~\cite{ycb} dataset. It contains 66K training images and 11K test images. We consider only objects seen in the training set for evaluation.

\paragraph{Metrics. }
\vspace{-12px}
As in \cite{hampali2020honnotate}, we report the mean joint error after scale-translation alignment of the root joint and the area-under-the-curve~(AUC) metrics to evaluate the hand pose.  The object pose is computed w.r.t to the hand frame of reference and is evaluated using the standard Maximum Symmetry-Aware Surface Distance (MSSD)~\cite{HodanSDLBMRM20}, as it considers the symmetricity of objects.

% and we provide details about the angles and axes of symmetry for different objects in HO-3D in the suppl. mat.% The MSSD metric is given by %$\text{err}_\text{mssd}  = \min_{R\in \mathcal{S}} \max_{\boldsymbol{x}\in V} \lvert \lvert \hat{P} \cdot \boldsymbol{x} - P^* \cdot R\cdot \boldsymbol{x}\lvert \lvert_2$,
 
\vspace{-12px}
\paragraph{Results. }
We use 3D joint representation to estimate the hand pose. Table~\ref{tab:ho3d_hand} compares the accuracy of the proposed hand-object pose estimation method with other approaches. Keypoint Transformer performs significantly better than previous methods~\cite{hampali2020honnotate, hasson19_obman, hasson20_handobjectconsist}. As \cite{hampali2020honnotate, hasson19_obman, hasson20_handobjectconsist} do not consider symmetricity of objects during training and evaluation, we also report our results in a similar setting. %We also use a ResNet-18 backbone in this experiment as in \cite{hasson19_obman, hasson20_handobjectconsist}. 
We show qualitative results in Fig.~\ref{fig:ho_qual}. Please refer to supplementary material for quantitative comparison with more recent works.

% Table~\ref{tab:ho3d_obj} compares the accuracy of the estimated object poses with \cite{hasson20_handobjectconsist}. As \cite{hasson20_handobjectconsist} does not consider symmetricity of objects, we show results with ('Ours-sym') and without ('Ours-nosym') considering symmetry in our training. Our method obtains more accurate hand-relative object poses. We show some qualitative results in Fig.~\ref{fig:ho_qual} and suppl. mat.

\begin{table}
      \centering
      \scalebox{.8}{
        \begin{tabular}{@{}l P{1.2cm} P{1.0cm} P{1.0cm} P{1.8cm} P{2.5cm}@{}}
            \toprule
            & \multirow{2}{*}{\parbox{1.4cm}{\centering Camera Intrinsics}} & 
            \multirow{2}{*}{\parbox{1.0cm}{\centering Image Crop}} & 
            \multirow{2}{*}{\parbox{1.0cm}{\centering Joint Error}} &
            \multirow{2}{*}{\parbox{1.8cm}{\centering Mean Joint AUC }} &
            \multirow{2}{*}{\parbox{2.5cm}{\centering MSSD (Object Pose Error) }}\\
            &&&& \\
            \midrule
            \cite{hampali2020honnotate} & Yes & Yes & 3.04 & 0.49 & - \\
            \cite{hasson19_obman} & No & Yes & 3.18 & 0.46 & - \\
            \cite{hasson20_handobjectconsist} & Yes & No & 3.69 & 0.37 & 11.99 \\
            Ours & No & Yes & \textbf{2.57} & \textbf{0.54} & \textbf{7.02} \\
            \bottomrule
            \end{tabular}}
            \vspace{-0.2cm}
            \caption{\textbf{Accuracy of our method on the HO-3D dataset for hand and object pose estimation.} Our method outperforms previous methods by a large margin.}
            \vspace{-0.5cm}
            % \caption{{\bf Hand 3D joint accuracy evaluated on HO-3D~\cite{hampali2020honnotate}.} Our transformer-based architecture performs significantly better than existing approaches that directly predict hand model parameters~\cite{hasson19_obman,hasson20_handobjectconsist} or predict hand keypoints~\cite{hampali2020honnotate}.}
    \label{tab:ho3d_hand}
\end{table}%

\subsection{\textbf{$\twohodataset$}}

\paragraph{Training and test sets. }
We introduce a dataset named $\twohodataset$ comprising sequences of two hands manipulating an object automatically annotated with the 3D poses of the hands and the object, by extending the work of \cite{hampali2020honnotate} to consider two hands. Figs.~\ref{fig:teaser} and \ref{fig:ho_qual} show some images. Five different subjects manipulate 10 different objects from YCB using both hands with a functional intent. We captured 60'998 training images and 15'342 test images using a 5 RGBD cameras multi-view setup. The $\twohodataset$ test set contains 7 objects seen in the training set and 1 unseen object.  More details are provided in the supplementary material. $\twohodataset$ is significantly more challenging than previous hand interaction datasets as there are many large occlusions between the hands and the objects.

\vspace{-12px}
\paragraph{Metrics and Results}
We use the 3D joint representation for the hand pose and evaluate the accuracy using the MPJPE and MRRPE metrics~(see Section~\ref{sec:interhands}) for the hand and the MSSD metric for the object~(see Section~\ref{sec:ho3d}). Details about the angle of symmetry for different objects considered during training and evaluation is provided in the suppl. mat. Due to large occlusions of the object by the hands, a portion of images are unsuitable for object pose estimation. We identify these images as the ones whose ground truth object segmentation area is less than 2\% of the cropped image area and exclude them from the object pose estimation during training and evaluation.
We also used the HO-3D train split and mirrored the images randomly during training to obtain right hand- and left hand-only images, to later combine with the training set of $\twohodataset$.

Our method achieves an MPJPE of 3.09 cm and an MRRPE of 8.28 cm on this dataset. Due to large occlusions by the object, estimating the translation between the hands is more challenging and the MRRPE is about 2.5 times worse than on InterHands2.6M which does not contain objects. On objects, our method achieves MSSD values of 7.96~cm. We provide object-specific MSSD values in the supplementary material.  Fig.~\ref{fig:ho_qual} shows qualitative results.

% \vspace{-5px}
\section{Discussion}
\label{sec:discussions}

\begin{figure}
\begin{center}
\includegraphics[trim=160 160 200 150,clip, width=1\linewidth]{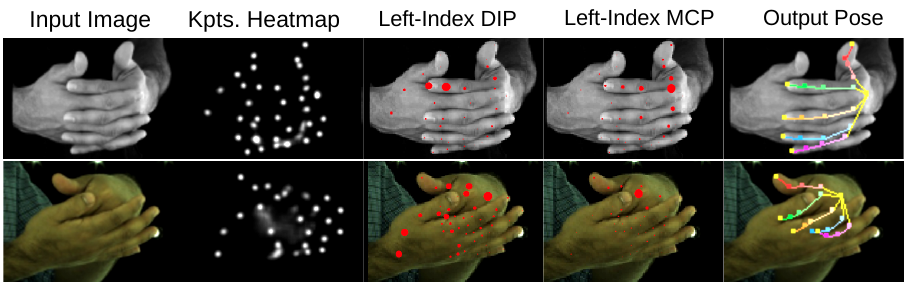} 
\end{center}
\vspace{-0.6cm}
\caption{{\bf Visualizing the cross-attention weights for two joint queries of the left hand.} The radius of the red circles are proportional to the weights. When the joint is occluded like the DIP joint on the second row, nearby visible keypoints are selected by the attention mechanism for pose estimation.}
\label{fig:attn}
\vspace{-0.3cm}
\end{figure}

\begin{figure}
\begin{center}
\includegraphics[trim=135 135 100 85,clip, width=0.98\linewidth]{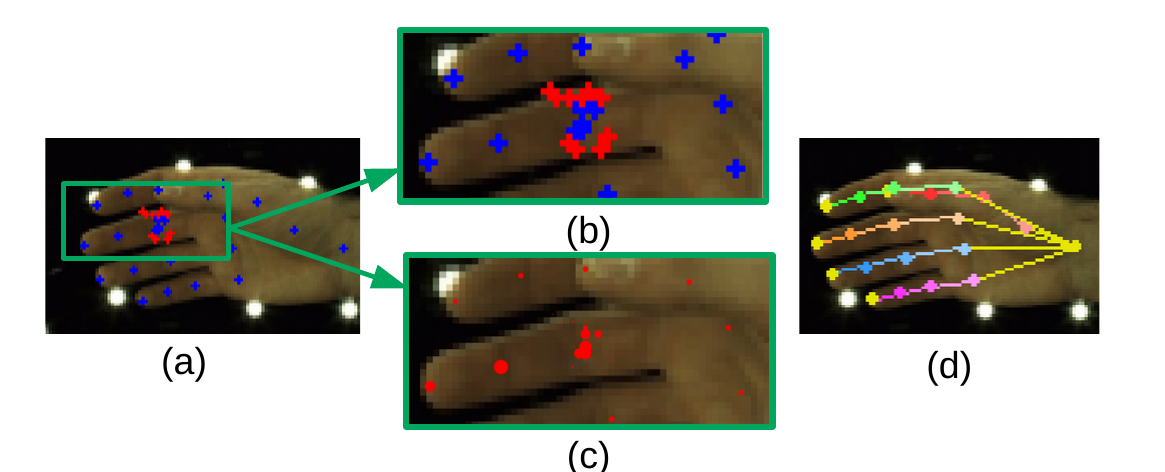}   
\end{center}
\vspace{-0.6cm}
\caption{\textbf{Robustness to noisy keypoints.} In this example, we added noisy keypoints around the middle finger PIP joint. Most of the noisy keypoints are predicted to belong to background class~(in red), while some are associated with the PIP joint~(in blue). The noisy keypoints associated with the PIP joint have all higher cross-attention weights (c) and are considered for final pose estimation.}
\label{fig:robust}
\vspace{-0.3cm}
\end{figure}

\begin{figure}[]
\begin{center}
\includegraphics[trim=185 230 250 80,clip, width=0.99\linewidth]{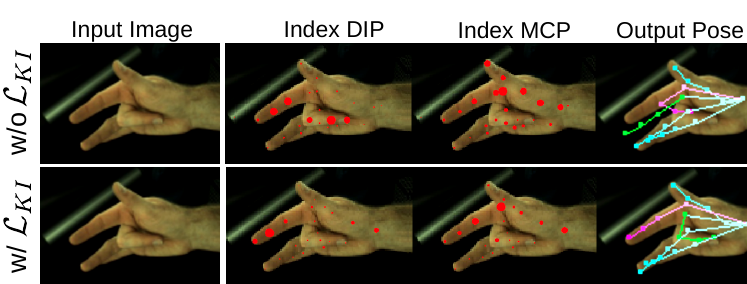} 
\end{center}
\vspace{-0.6cm}
\caption{\textbf{Cross-attention with and without the keypoint-joint association loss $\calL_\text{KI}$}. $\calL_\text{KI}$ makes the keypoints `identity-aware', resulting in higher accuracy.}
% For each joint query, the decoder selects the image features only from the location of the joint and estimates joint-related pose parameters such as joint angle and joint vector.}
% \sayanrmk{TODO: Change Font Of Fig}}
\label{fig:ki_comp}
\vspace{-0.3cm}
\end{figure}

% For each joint query, the decoder selects the image features only from the location of the joint and estimates joint-related pose parameters such as joint angle and joint vector.}
% \sayanrmk{TODO: Change Font Of Fig}}

We report here the results of experiments we perform using InterHand2.6M (V0.0) to understand better our method. %More details about these experiments are provided in the suppl. mat.

\vspace{-0.5cm}

\paragraph{Visualization of cross-attention.} We visualize the cross-attention weights in Fig.~\ref{fig:attn} for two joint queries of the left-index finger. When the joint is not occluded, as in the first row, each joint query attends to the keypoint whose location coincides with the corresponding joint location in the image. In other words, local image features at the location of the joint are used to estimate the pose of that joint. We believe this property of using local image features helps in achieving higher accuracy than other CNN-based approaches~\cite{Moon_2020_ECCV_InterHand2.6M, hasson19_obman, hasson20_handobjectconsist}.
In the second row of Fig.~\ref{fig:attn}, the left index finger is occluded except for the MCP joint and no keypoints 
are detected for the invisible joints. The cross-attention module \textit{selects} nearby visible keypoints resulting in a more global-level feature for estimating the joint pose.

%When the joint becomes visible such as the left-index MCP joint, the local image features are used.

\vspace{-0.5cm}

\paragraph{Robustness to noisy keypoints.} 
To demonstrate the robustness of our method, we added incorrect keypoints around the detected keypoints. As shown in Fig.~\ref{fig:robust} and supplementary material, most of these keypoints are labeled as background and all the keypoints that are assigned to the same joint are considered equally for the pose estimation.

\vspace{-0.5cm}

\paragraph{Importance of the keypoint-joint association loss $\calL_\text{KI}$.} $\calL_\text{KI}$ helps the cross-attention module to select the appropriate features for pose estimation as visualized in Fig.~\ref{fig:ki_comp}. Further, $\calL_\text{KI}$ improves MPJPE by 10\% (17.08 mm v/s 18.91 mm) and MRRPE by 15\% (33.14 mm v/s 38.96 mm) on interacting hand images.

% \paragraph{Importance of the keypoint identity loss.} Use of keypoint-joint association loss, $\calL_\text{KI}$ during training helps the cross-attention module to select the approriate features for pose estimation as visualized in Fig.~\ref{fig:ki_comp}. Further, $\calL_\text{KI}$ improves MPJPE by 10\% (17.08mm v/s 18.91mm) and MRRPE by 15\% (33.14mm v/s 38.96mm) on interacting hand images from InterHand2.6M dataset.

\vspace{-0.5cm}

\paragraph{Accuracy with different pose representations.} Table~\ref{tab:pose_rep_comp} compares the accuracy of the 3 hand pose representations that we consider. While the accuracy of the 3D and 2.5D representations are similar, the joint angle representation results in lower accuracy, in line with the observation from previous works~\cite{Pavlakos2018LearningTE, Pavlakos2019TexturePoseSH, hasson19_obman, hasson20_handobjectconsist, Angjoo}. 

%This observation is in line with previous works~\cite{Pavlakos2018LearningTE, Pavlakos2019TexturePoseSH, hasson19_obman, hasson20_handobjectconsist, Angjoo} that estimate the MANO joint angles or their PCA components, resulting in higher mean joint error compared to the 2.5D representation.

\begin{table}
    \centering
    \scalebox{0.7}{
    \begin{tabular}{@{}cccccc@{}}
    % \begin{tabular}{@{}P{2.0cm} P{1.5cm} P{1.8cm} P{1.6cm} P{1.0cm} P{1.8cm}@{}}
    \toprule
    & \multirow{2}{*}{\parbox{1.4cm}{\centering Camera Intrinsics}} & \multicolumn{3}{c}{\centering MPJPE (mm)} &  \multirow{2}{*}{\parbox{1.5cm}{\centering MRRPE (mm)}}  \\
    & & Single Hand & Two Hands & All &  \\
    \midrule
    % InterNet \cite{Moon_2020_ECCV_InterHand2.6M} & Yes & 13.79 & 21.24 & 17.54 & 40.46 \\
    % \hline
    %%% These numbers are with V0.0 machine_annot
    % $\theta$* & No & 14.00 & 19.16 & 16.61 & 37.91 \\
    3D & No & 12.42 & \textbf{17.08} & 14.76 & \textbf{33.14} \\
    2.5D & Yes & \textbf{11.73} & 17.69 & \textbf{14.73} & 34.40 \\
    $\theta$ & No & 15.36 & 20.61 & 18.01 & 37.91 \\
    \bottomrule
    \end{tabular}
    }
    \vspace{-0.2cm}
    \caption{\textbf{Accuracy obtained with the 3 different pose representations.}}
    \label{tab:pose_rep_comp}
    \vspace{-0.4cm}
\end{table}

\begin{table}
    \centering
    \scalebox{0.75}{
    \begin{tabular}{@{}l P{0.75cm} P{0.75cm} P{0.75cm} |  P{0.75cm} P{0.75cm} P{0.75cm} | P{0.75cm}@{}}
    % \begin{tabular}{@{}P{1.75cm}P{1.2cm} P{1.2cm} P{1.2cm} |  P{1.2cm} P{1.2cm} P{1.2cm} | P{0.75cm}@{}}
    \toprule
    & \multicolumn{3}{c |}{\centering $N_{CA}=1$, Varying $N_{SA}$} &  \multicolumn{3}{c | }{\centering $N_{SA}=6$, Varying $N_{CA}$} & \multirow{2}{*}{\parbox{0.75cm}{\centering \cite{Moon_2020_ECCV_InterHand2.6M}}}  \\
    & $0$ & $3$ & $6$ & $1$ & $3$ & $6$ &  \\
    \midrule
    %%% These numbers are with V0.0, all
    Single Hand & 12.34 & 11.77 & 11.24 & 11.24 & 11.14 & 11.08 & 12.63 \\
    Two Hands & 16.93 & 15.55 & 15.44 & 15.44 & 15.35 & 15.33 & 17.36 \\
    % InterNet \cite{Moon_2020_ECCV_InterHand2.6M} & Yes & 13.79 & 21.24 & 17.54 & 40.46 \\
    \bottomrule
    \end{tabular}
    }
    \vspace{-0.2cm}
    \caption{\textbf{3D pose accuracy~(MPJPE, in mm) for different numbers of self-attention~($N_{SA}$) and cross-attention~($N_{CA}$) layers.}}
    \label{tab:enc_dec_layers}
    \vspace{-0.2cm}
\end{table}

\begin{table}
    \centering
    \scalebox{0.73}{
    \begin{tabular}{@{}l P{1.5cm} P{1.6cm} P{1.6cm} P{2.6cm}@{}}
    \toprule
    % & \multicolumn{3}{c | }{\centering $L_{D}=1$} &  \multicolumn{3}{c | }{\centering $L_{E}=6$} & \multirow{2}{*}{\parbox{0.75cm}{\centering \cite{Moon_2020_ECCV_InterHand2.6M}}}  \\
    & Resnet-18 & Resnet-34 & Resnet-50 & \cite{Moon_2020_ECCV_InterHand2.6M} (Resnet-50)  \\
    \midrule
    %%% These numbers are with V0.0, all
    Total Params & 28M & 38M & 48M & 48M\\
    \midrule
    Single Hand & 11.67 & 11.99 & 11.28 & 12.63 \\
    Two Hands & 16.78 & 16.41 & 15.32 & 17.36 \\
    % InterNet \cite{Moon_2020_ECCV_InterHand2.6M} & Yes & 13.79 & 21.24 & 17.54 & 40.46 \\
    % \hline
    \bottomrule
    \end{tabular}
    }
    \vspace{-0.2cm}
    \caption{\textbf{3D pose accuracy~(MPJPE, in mm) for different backbones. }}
    \label{tab:num_params}
    \vspace{-0.4cm}
\end{table}

\vspace{-0.5cm}

\paragraph{Effect of the number of self-attention (SA) and cross-attention (CA) layers.}
Table~\ref{tab:enc_dec_layers} reports the MPJPE  with different combinations of SA and CA layers. Even in the absence of any SA layers, our method outperforms \cite{Moon_2020_ECCV_InterHand2.6M}. Adding more CA layers has little effect on the accuracy.

\vspace{-0.5cm}

\paragraph{Effect of the number of parameters.}
Table~\ref{tab:num_params} reports the MPJPE for different CNN backbones. While larger backbones improve the accuracy, our method outperforms \cite{Moon_2020_ECCV_InterHand2.6M} even with a Resnet-18 backbone with approximately half the total number of parameters.

\vspace{-5px}
\section{Conclusion}
\label{sec:conclusion}
\vspace{-5px}
We showed that, by integrating a keypoint detector into a Transformer architecture, we could predict  3D poses of hands and objects  from very challenging images, in a much more accurate way than a standard Transformer architecture does. As we rely on keypoints, we believe that our approach is more general and could be applied to other problems, such as human and other articulated objects pose prediction and object category pose prediction~\cite{pavlakos17object3d}.
\vspace{-10px}
\paragraph{Acknowledgments.} This work was supported by the Christian Doppler Laboratory for Semantic 3D Computer Vision, funded in part by Qualcomm Inc, and Chistera IPalm.

%%%%%%%%%Remove this if not submitting to arxiv%%%%%%%%%%%%%%%%%
\maketitle
\setcounter{section}{0}% Reset numbering for sections
\renewcommand{\thesection}{\Alph{section}}% Adjust section printing (from here onward)
\begin{center}
\section*{Supplementary Material}    
\end{center}

In this supplementary material, we discuss the limitations of our method, provide more details about the experiments and also show several qualitative results and comparisons. We also refer the reader to the \textbf{Supplementary Video} for visualization of results on different action sequences.

\section{Hand Pose Representations and Losses}
\label{sec:pose_rep_loss}
% \begin{figure}
%     \centering
%     \includegraphics[width=0.5\linewidth]{figures/supplementary/hand_joints.png}
%     \caption{Samples from $\twohodataset$ dataset. Our dataset contains sequences with complex actions performed by both hands on YCB~\cite{ycb} objects.}
%     \label{fig:hand_joints}
% \end{figure}

We detail the three possible representations mentioned in Section~3.4 of the paper. We assume 21 3D-joint locations per hand as in the MANO~\cite{Romero2017EmbodiedH} model. The losses for each of the 3 representations are summarized in Table~\ref{tab:pose_loss}.

\vspace{-12px}

\paragraph{3D representation.}  In this representation, each joint $j$ is associated with a parent-relative joint vector $V(j) = J_\ThD(j) - J_\ThD(p(j))$, where $J_\ThD$ is the 3D joint location and $p(j)$ refers to the parent joint index of joint $j$. We estimate 20 joint vectors per hand using 20 joint queries, one for each skeletal bone (40 queries for two hands), from which we can compute the root-relative 3D location, $J_{3D}^r$ of each joint by simple  accumulation. The advantage of this representation is that it defines the hand pose relative to its root without requiring knowledge of the camera intrinsics.

\vspace{-12px}

\paragraph{2.5D representation~\cite{umar, Moon_2020_ECCV_InterHand2.6M}.} 

In this representation, each joint is parameterised by its 2D location $J_\TwD$, and the difference $\Delta Z^p$ between its  depth and the depth of its parent joint. The camera intrinsics matrix $K$ and the absolute depth $Z_\text{root}$ of the root joint (the wrist)~\cite{Moon_2020_ECCV_InterHand2.6M} or the scale of the hand~\cite{umar} are then required to reconstruct the 3D pose of the hand in camera coordinate system as $J_\ThD = K^{-1} \cdot (Z_\text{root}+\Delta Z^r) \cdot  \big[J_{\TwD_x},J_{\TwD_y},1\big]^T \>,$
where $\Delta Z^r$ is the root-relative depth of the joint computed from its predicted $\Delta Z^p$ and the predicted $\Delta Z^p$ for its parents. $J_{\TwD_x}, J_{\TwD_y}$ are the predicted $x$ and $y$ coordinates of $J_\TwD$. 

When using this representation, we also predict the root depth $Z_\text{root}$ separately using RootNet~\cite{Moon2019CameraDT} as in \cite{Moon_2020_ECCV_InterHand2.6M}. Each joint query estimates the $J_{\TwD}$ and $\Delta Z^r$ for that joint and we require a total of 21 joint queries (42 for two hands), one for each joint location to estimate the 2.5D pose per hand. 

\vspace{-12px}

\paragraph{MANO joint angles, $\theta$~\cite{Romero2017EmbodiedH}.} 
In this representation, each 3D hand pose is represented by 16 3D joint angles in the hand kinematic tree and is estimated using 16 joint queries per hand, one for each joint. The MANO hand shape parameter is estimated along with the relative translation between the hands using an additional query. Given the predicted 3D joint angles $\theta$ for each hand and the shape parameters $\beta$, it is possible to compute the root-relative 3D joint locations, $J_{3D}^r$ of each hand.

\begin{table}
    \centering
    \scalebox{0.8}{
    \begin{tabular}{@{}P{2.0cm}P{8.0cm}@{}}
    \toprule
    Representation & $\calL_{hand-pose}$ \\
    \midrule
    3D & $\sum_j \lvert\lvert V(j) - V(j)^*\lvert\lvert_1 + \sum_j \lvert\lvert J_{3D}^r(j) - J_{3D}^{r^*}(j)\lvert\lvert_1$\\
    2.5D & $\sum_j \lvert\lvert J_{2D}(j) - J_{2D}^*(j)\lvert\lvert_1 + \sum_j \lvert\Delta Z^r(j) - \Delta Z^{r^*}(j) \lvert$\\
    $\theta$ & $\sum_j \lvert\lvert J_{3D}^r(j) - J_{3D}^{r^*}(j)\lvert\lvert_1 + \sum_j \lvert\lvert \theta (j) - \theta^* (j) \lvert\lvert_1$\\
    \bottomrule
    \end{tabular}
    }
    \vspace{-5px}
    \caption{Hand pose losses for different pose representations. $x^*$ denotes the ground-truth values for variable $x$ and $x(j)$ the value of $x$ at joint $j$.
}
    \vspace{-12px}
    \label{tab:pose_loss}
\end{table}

\section{Comparison with state-of-the-art on HO3D Dataset}
We compare the performance our method with several other methods on the HO-3D(V2) and HO-3D(V3) datasets and show the results in Tab.~\ref{tab:ho3d_v2_comp} and Tab.~\ref{tab:ho3d_v3_comp}, respectively. On HO-3D(V2), the performance of our method is very close to the HandOccNet~\cite{handoccnet}, which achieves the highest accuracy when considering the scale-translation aligned MPJPE metric. However, HandOccNet achieves higher accuracy when considering the procrustes aligned MPJPE metric. On the HO-3D(V3) dataset, our method performs worse than ArtiBoost~\cite{artiboost}, which uses additional training data.
\begin{table*}
    \centering
    \scalebox{0.7}{
    \begin{tabular}{@{}P{2.8cm}P{1.7cm}P{2.3cm}P{2.0cm}P{2.0cm}P{2.0cm}P{2.0cm}P{2.0cm}P{2.0cm}P{2.0cm}@{}}
    \toprule
    \multirow{2}{*}{\parbox{2.5cm}{\centering Method}} &
    \multirow{2}{*}{\parbox{1.7cm}{\centering Pose Repr.}} &
    \multirow{2}{*}{\parbox{2.3cm}{\centering Joint Error (scale and trans. align.) in cms}} & \multirow{2}{*}{\parbox{2.5cm}{\centering Joint Error AUC (scale and trans. align.)}} & \multirow{2}{*}{\parbox{2.5cm}{\centering Joint Error (Procrustes align.) in cms}} & 
    \multirow{2}{*}{\parbox{2.5cm}{\centering Joint Error AUC (Procrustes align.)}} &
    \multirow{2}{*}{\parbox{2.5cm}{\centering Mesh Error (Procrustes align.) in cms}} &
    \multirow{2}{*}{\parbox{2.5cm}{\centering Mesh Error AUC (Procrustes align.)}} &
    \multirow{2}{*}{\parbox{2.0cm}{\centering Mesh Error F-Score @5mm}} &
    \multirow{2}{*}{\parbox{2.0cm}{\centering Mesh Error F-Score @15mm}}\\ \\ \\
    \midrule
    METRO~\cite{metro} & Mesh & 2.89 & 0.504 & 1.04 & 0.792 & 1.11 & 0.779 & 0.484 & 0.946 \\
    Liu et al.~\cite{liu} & Joint angle & 3.17 & 0.463 & 0.99 & 0.803 & 0.95 & 0.810 & 0.528 & 0.956 \\
    HandOccNet~\cite{handoccnet} & Joint angle & 2.40 & 0.557 & 0.91 & 0.819 & 0.88 & 0.819 & 0.564 & 0.963 \\
    I2L-MeshNet~\cite{Moon2020I2LMeshNetIP} & Mesh & 2.60 & 0.529 & 1.12 & 0.775 & 1.39 & 0.722 & 0.409 & 0.932 \\
    Pose2Mesh~\cite{pose2mesh} & Mesh & 3.33 & 0.480 & 1.25 & 0.754 & 1.27 & 0.749 & 0.441 & 0.909 \\
    I2UV-HandNet~\cite{i2v} & Mesh & - &- & 0.99 & 0.804 & 1.01 & 0.799 & 0.500 & 0.943 \\
    Zheng et al.~\cite{zheng} & 2.5D & 2.51 & 0.541 & - & - & - & - & - & - \\
    Hampali et al.~\cite{hampali2020honnotate} & 3D & 3.04 & 0.494 & 1.07 & 0.788 & 1.06 & 0.790 & 0.506 & 0.942 \\
    Hasson et al.~\cite{hasson19_obman} & Joint angle & 3.18 & 0.461 & 1.10 & 0.780 & 1.12 & 0.777 & 0.464 & 0.939 \\
    Hasson et al.~\cite{hasson20_handobjectconsist} & Joint angle & 3.69 & 0.369 & 1.14 & 0.773 & 1.14 & 0.773 & 0.428 & 0.932 \\ 
    ArtiBoost~\cite{artiboost} & 2.5D & 2.53 & 0.532  & 1.14 & 0.773 & 1.09 & 0.782 & 0.488 & 0.944 \\
    \midrule
    Ours-ResNet18 & Joint angle & 3.11 & 0.459 & 1.10 & 0.780 & 1.13 & 0.774 & 0.444 & 0.935\\
    Ours-ResNet50 & 3D & 2.57 & 0.553 & 1.08 & 0.786 & - & - & - & -\\
    
    % 3D & $\sum_j \lvert\lvert V(j) - V(j)^*\lvert\lvert_1 + \sum_j \lvert\lvert J_{3D}^r(j) - J_{3D}^{r^*}(j)\lvert\lvert_1$\\
    % 2.5D & $\sum_j \lvert\lvert J_{2D}(j) - J_{2D}^*(j)\lvert\lvert_1 + \sum_j \lvert\Delta Z^r(j) - \Delta Z^{r^*}(j) \lvert$\\
    % $\theta$ & $\sum_j \lvert\lvert J_{3D}^r(j) - J_{3D}^{r^*}(j)\lvert\lvert_1 + \sum_j \lvert\lvert \theta (j) - \theta^* (j) \lvert\lvert_1$\\
    \bottomrule
    \end{tabular}
    }
    \vspace{-5px}
    \caption{Comparison with state-of-the-art methods on HO-3D V2 dataset.}
    % \vspace{-12px}s
    \label{tab:ho3d_v2_comp}
\end{table*}

\begin{table*}
    \centering
    \scalebox{0.7}{
    \begin{tabular}{@{}P{2.8cm}P{1.7cm}P{2.3cm}P{2.0cm}P{2.0cm}P{2.0cm}P{2.0cm}P{2.0cm}P{2.0cm}P{2.0cm}@{}}
    \toprule
    \multirow{2}{*}{\parbox{2.5cm}{\centering Method}} &
    \multirow{2}{*}{\parbox{1.7cm}{\centering Pose Repr.}} &
    \multirow{2}{*}{\parbox{2.3cm}{\centering Joint Error (scale and trans. align.) in cms}} & \multirow{2}{*}{\parbox{2.5cm}{\centering Joint Error AUC (scale and trans. align.)}} & \multirow{2}{*}{\parbox{2.5cm}{\centering Joint Error (Procrustes align.) in cms}} & 
    \multirow{2}{*}{\parbox{2.5cm}{\centering Joint Error AUC (Procrustes align.)}} &
    \multirow{2}{*}{\parbox{2.5cm}{\centering Mesh Error (Procrustes align.) in cms}} &
    \multirow{2}{*}{\parbox{2.5cm}{\centering Mesh Error AUC (Procrustes align.)}} &
    \multirow{2}{*}{\parbox{2.0cm}{\centering Mesh Error F-Score @5mm}} &
    \multirow{2}{*}{\parbox{2.0cm}{\centering Mesh Error F-Score @15mm}}\\ \\ \\
    \midrule
    ArtiBoost~\cite{artiboost} & 2.5D & 2.34 & 0.565  & 1.08 & 0.785 & 1.04 & 0.792 & 0.507 & 0.946 \\
    % \midrule
    Ours-ResNet50 & 3D & 2.48 & 0.575 & 1.09 & 0.785 & - & - & - & -\\
    
    % 3D & $\sum_j \lvert\lvert V(j) - V(j)^*\lvert\lvert_1 + \sum_j \lvert\lvert J_{3D}^r(j) - J_{3D}^{r^*}(j)\lvert\lvert_1$\\
    % 2.5D & $\sum_j \lvert\lvert J_{2D}(j) - J_{2D}^*(j)\lvert\lvert_1 + \sum_j \lvert\Delta Z^r(j) - \Delta Z^{r^*}(j) \lvert$\\
    % $\theta$ & $\sum_j \lvert\lvert J_{3D}^r(j) - J_{3D}^{r^*}(j)\lvert\lvert_1 + \sum_j \lvert\lvert \theta (j) - \theta^* (j) \lvert\lvert_1$\\
    \bottomrule
    \end{tabular}
    }
    \vspace{-5px}
    \caption{Comparison with state-of-the-art methods on HO-3D V3 dataset. Note that ArtiBoost~\cite{artiboost} uses additional training data which is not used in our method.}
    \vspace{-12px}
    \label{tab:ho3d_v3_comp}
\end{table*}

\section{Method Limitations}
Though our method results in accurate poses during interactions, the results are sometimes not plausible as we do not model contacts and interpenetration~\cite{Karunratanakul2020GraspingFL,Brahmbhatt2020ContactPoseAD,hasson19_obman} between hands and objects. Further, during highly complex and severely occluded hand interactions as we show in the last row of Fig.~\ref{fig:inter_qualitative}, our method fails to obtain reasonable hand poses. We believe these problems can be tackled in the future by incorporating temporal information and physical modeling into our architecture.

\section{Hand-Object Pose Estimation Pipeline}
\begin{figure*}[h]
\begin{center}
  \includegraphics[trim=10 100 30 34,clip, width=1\linewidth]{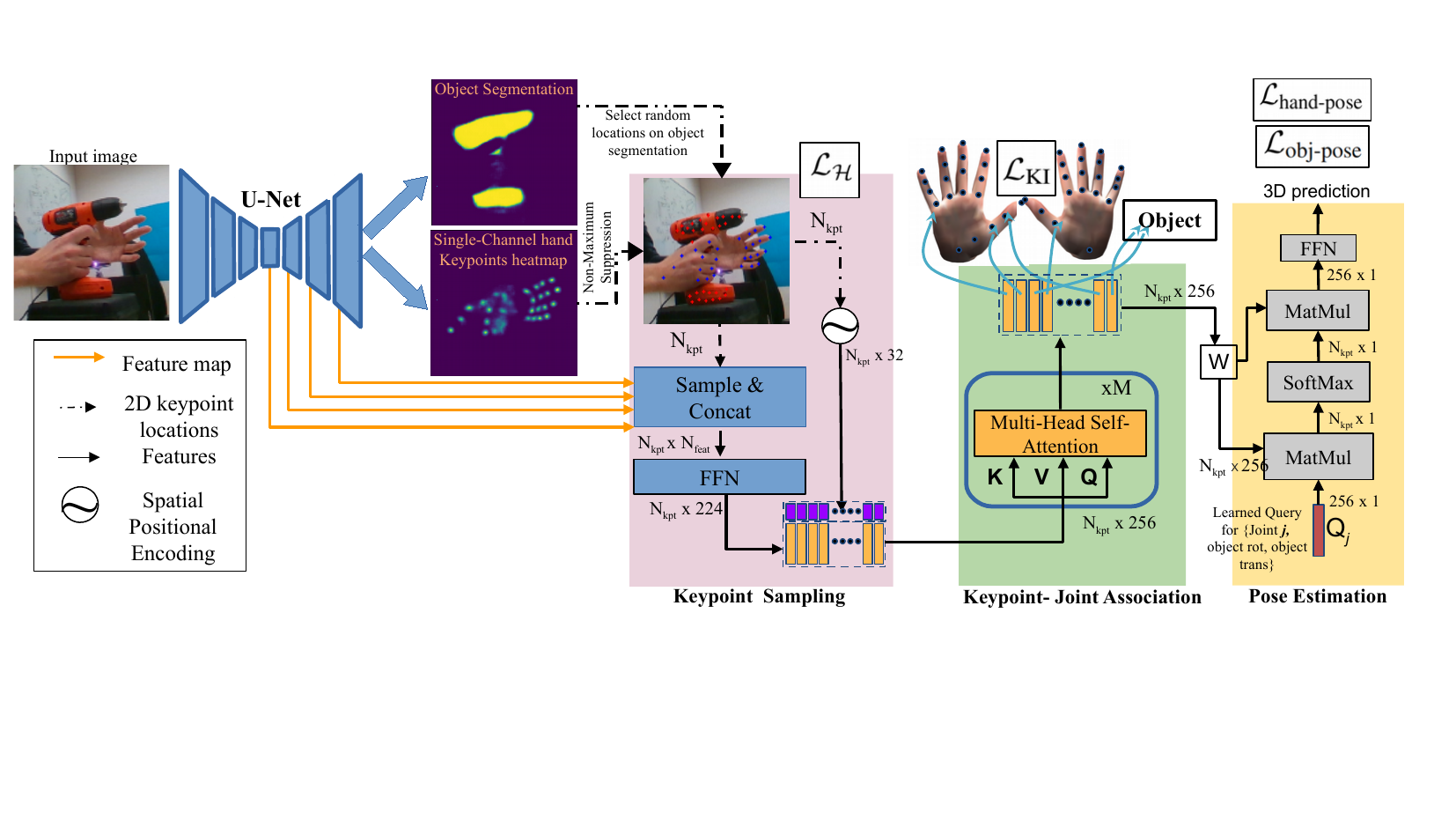} 
\end{center}
\vspace{-0.5cm}
\caption{{\bf Pipeline for hands and object pose estimation.} The object keypoints are selected by randomly sampling 2D locations on the object segmentation map regressed by the U-Net (Section 3.5 of main paper). The hand keypoints are selected from the single-channel keypoints heatmap, also regressed by the U-Net (Section 3.1 of main paper). Each of the detected keypoints are encoded using CNN image features and spatial embedding. The keypoints are associated with one of the 42 hand joints (21 joints per hand), the object class or the background class in the keypoint-joint association stage (Section 3.2 of main paper). The object rotation and translation w.r.t the right hand is estimated in the pose estimation stage using 2 different learned object queries, while the pose of each hand-joint is estimated using per-joint learned queries (Section 3.3 of main paper).}
\vspace{-0.5cm}
\label{fig:block_dig_w_obj}
\end{figure*}

In Fig.~\ref{fig:block_dig_w_obj}, we show the complete pipeline of our Keypoint-Transformer architecture for estimating poses of two hands and object during interaction.

\section{Implementation details}
The encoder of our U-Net~\cite{unet} is based on ResNet-50~\cite{He2016DeepRL} architecture while a series of upsampling and convolutional layers with skip connections forms the U-Net decoder. We use 256$\times$256 pixels as input image resolution, 128$\times$128 pixels as heatmap resolution, and set the 2D Gaussian kernel variance, $\sigma$ to 1.25 during training. The 256$\times$256 pixel input image patch is loosely cropped around the hand and object. We use Adam~\cite{KingmaB14} optimizer with a learning rate of $\text{10}^\text{-4}$ and $\text{10}^\text{-5}$ for the attention modules and CNN backbone, respectively. The network is trained for 50 epochs on 3 Titan V GPUs with a total batch size of 78 and uses on-line augmentation techniques such as rotation, scale and mirroring during training.

\section{Baseline Architectures}
\begin{figure}
\centering
    \begin{minipage}{1\linewidth}
        \centering
            \subfloat{\includegraphics[trim=100 210 80 100,clip,width=1\linewidth]{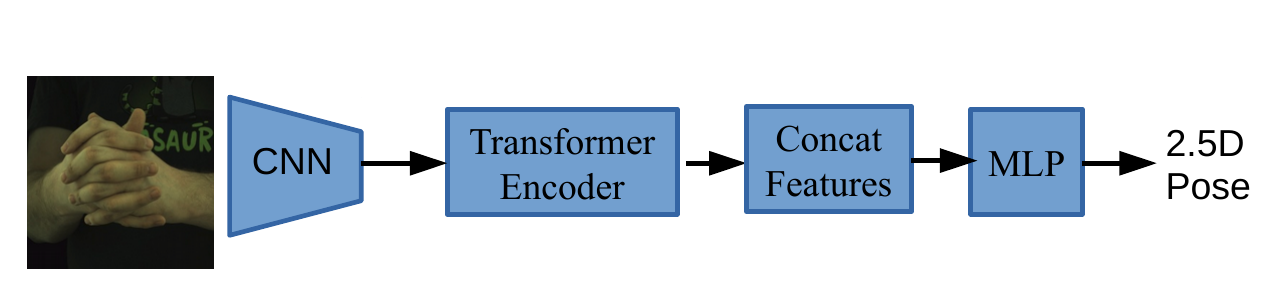}}
            \caption{The ‘CNN+SA’ baseline architecture.}
            \label{fig:baseline1}
    \end{minipage}%
\par\medskip
    \begin{minipage}{1\linewidth}
        \centering
            \subfloat{\includegraphics[trim=100 40 80 250,clip,width=1\linewidth]{figures/supplementary/baselines_block_dig.pdf}}
            \caption{The ‘CNN+SA+CA’ baseline architecture.}
            \label{fig:baseline2}
    \end{minipage}
    \vspace{-12px}
\end{figure}

We detail here the two baselines, ‘CNN+SA’ and ‘CNN+SA+CA’ considered in Section 4.1 of the main paper. Figures~\ref{fig:baseline1} and \ref{fig:baseline2} show their architectures. We used $256\times256$ cropped images as input to the CNN resulting in a feature map of spatial dimensions $8\times8$ and $2048$ channels. The features are flattened along the spatial dimensions and the 64 features are converted to 224 dimensions using 3 MLP layers. These features are then concatenated with 32-D positional embeddings resulting in 256-D features and are provided to the Transformer encoder. The networks were trained to output the 2.5D pose representation for 50 epochs on 3 Titan V GPUs with a batch size of 78. The joint queries in ‘CNN+SA+CA’ are learned in a similar way as for our Keypoint Transformer.

\section{Robustness to Noisy Keypoints}
\begin{figure}
    \centering
    \includegraphics[trim=100 10 170 10,clip, width=0.9\linewidth]{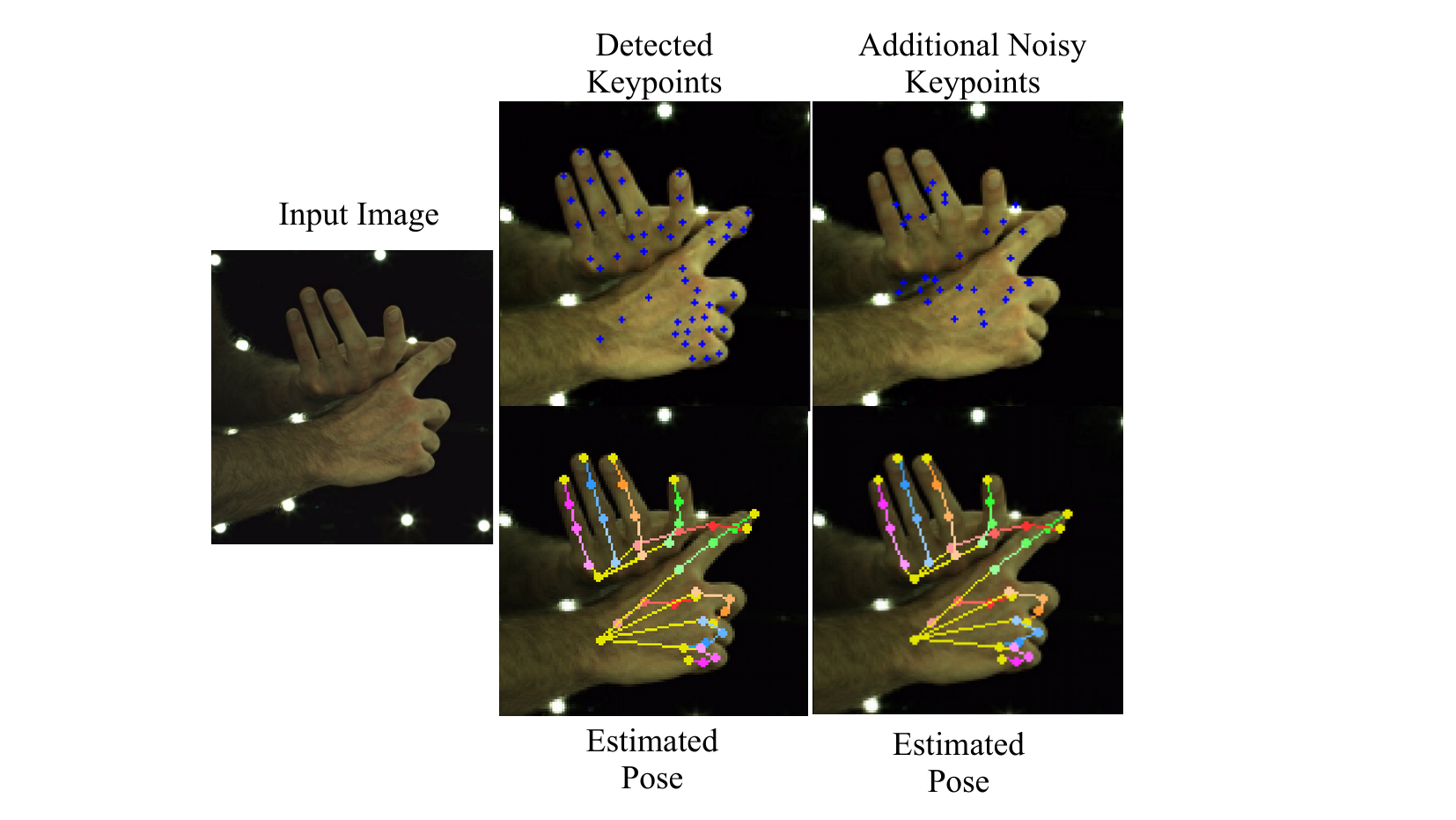}
    \caption{Effect of adding additional noisy keypoints. Our method predicts accurate poses even with noisy keypoints.}
    \label{fig:noise_added}
    \vspace{-10px}
\end{figure}

\begin{figure}
    \centering
    \includegraphics[trim=100 0 170 10,clip, width=0.9\linewidth]{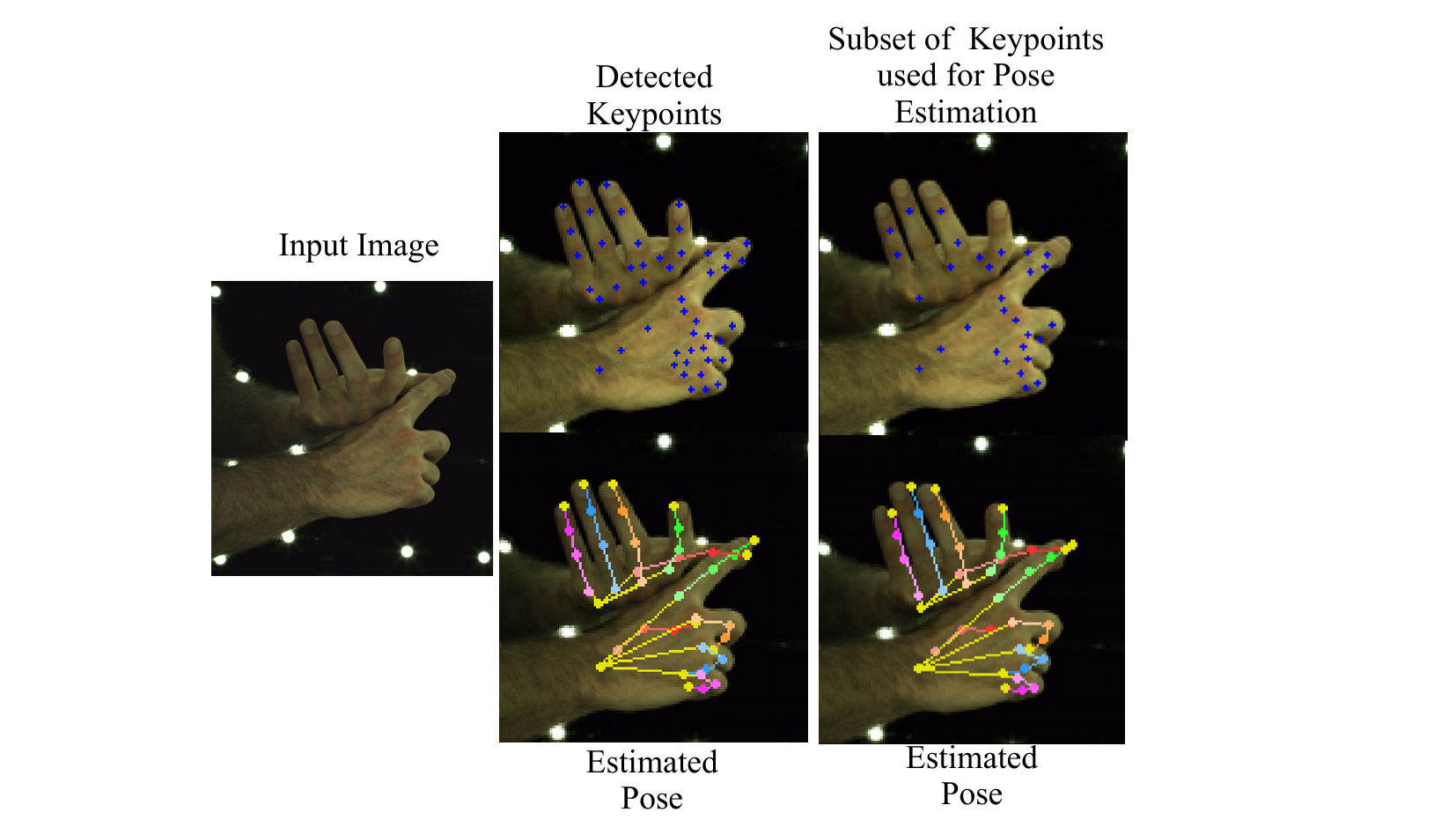}
    \caption{Effect of using a subset of detected keypoints for pose estimation. We consider only 30 of the 48 detected keypoints for pose estimation and still estimate an accurate pose.}
    \label{fig:noise_removed}
    \vspace{-12px}
\end{figure}

We show more examples to demonstrate the robustness of our method to noisy keypoints. We consider two scenarios, adding noisy keypoints to the set of detected keypoints, and randomly removing some keypoints from the set of detected keypoints. We show results in Figures~\ref{fig:noise_added} and \ref{fig:noise_removed}, respectively. The number of detected keypoints for these cases were 48 and we added 30 additional noisy keypoints for the former scenario and retained only 30 keypoints for the latter scenario.

%%%%%%%%%%%%%%%%%%%%%%%%%%%%%%%%%%%%%%%%%%%%%%%%%%%%%%%%%

%%%%%%%%%%%%%%%%%%%%%%%%%%%%%%%%%%%%%%%%%%%%%%%%%%%%%%%%%

\section{$\twohodataset$ Dataset}
\begin{figure*}
    \centering
    \includegraphics[trim=20 20 20 20,clip, width=1.0\linewidth]{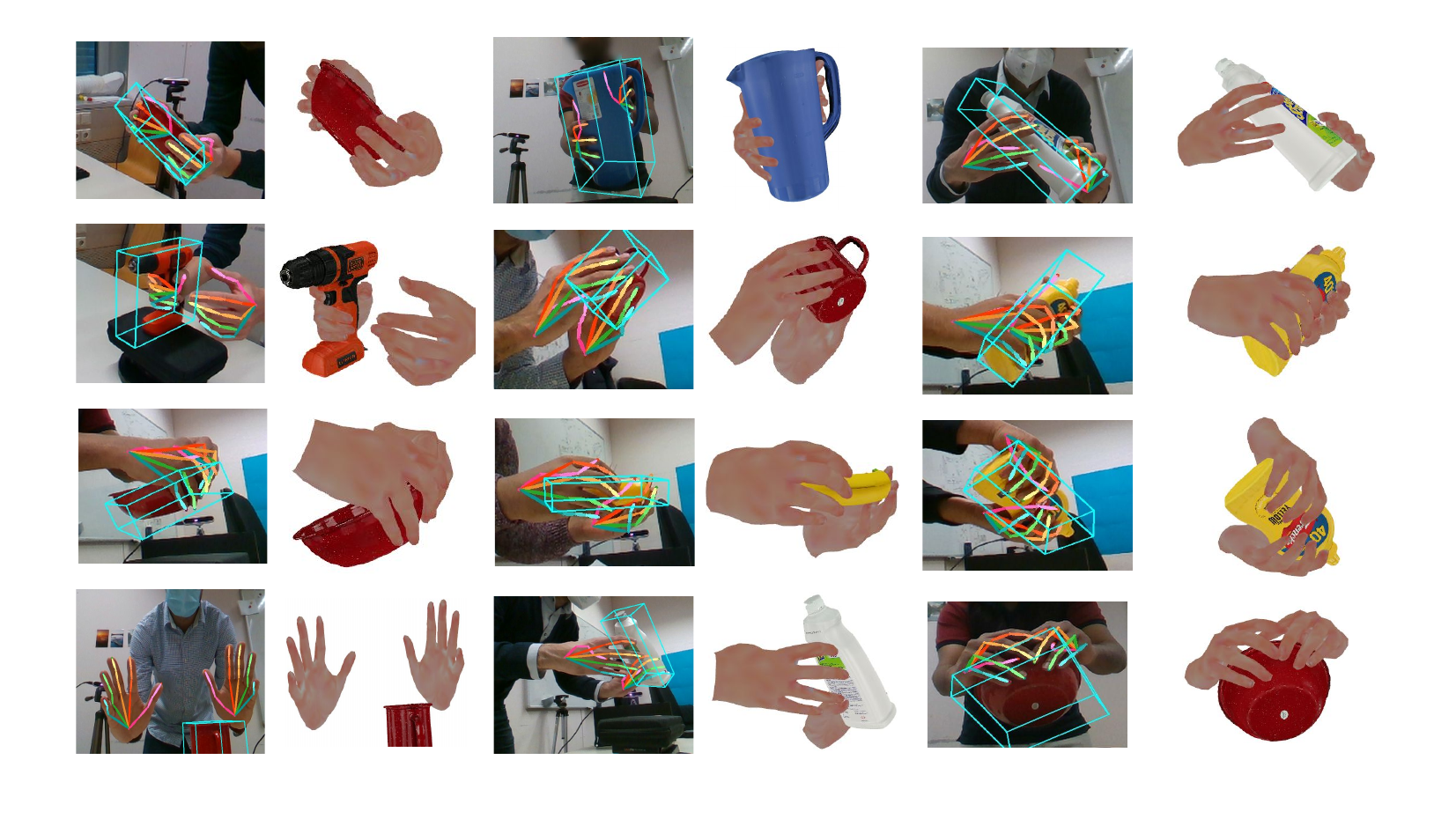}
    \caption{Samples from $\twohodataset$ dataset. Our dataset contains sequences with complex actions performed by both hands on YCB~\cite{ycb} objects.}
    \label{fig:h2o3d_samples}
\end{figure*}

Our dataset contains sequences of two hands interacting with an object, captured on a multi-view setup with 5 RGBD cameras. We collected data from six different subjects and considered ten objects from the YCB dataset with each subject manipulating the object with a functional intent. The dataset is automatically annotated with 3D poses of hands and objects using the optimization method of \cite{hampali2020honnotate}. The dataset  contains 60'998 training images and 15'342 test images from 17 different multi-view sequences in total. As explained in the main paper, we only consider 9'098 images from the set of 15'342 test images for object pose evaluation as the objects in the remaining images are barely visible due to occlusion by the hands.  We show some sample annotations from the dataset in Fig.~\ref{fig:h2o3d_samples}. Table~\ref{tab:sym} shows the list of YCB objects and their axis and angle of symmetry considered during our training and evaluation.

\subsection{Per-Object MSSD Values with Keypoint Transformer}
Table~\ref{tab:mssd_h2o3d} shows the accuracy of the object poses estimated by our Keypoint Transformer on the $\twohodataset$ dataset using the MSSD metric as described in Section~4.3 of the main paper.

\begin{table}[b]
    \centering
    \scalebox{1.0}{
    \begin{tabular}{P{2.5cm}|P{1.5cm}| P{1.5cm}}
    Object & Axis &  Angle  \\
    \hline
    Mustard Bottle & Z & $180^o$ \\
    Bleach Cleanser & Z & $180^o$ \\
    Cracker Box & Z & $180^o$ \\
    Sugar Box & Z & $180^o$ \\
    Potted Meat Can & Z & $180^o$ \\
    Bowl & Z & $\infty$ \\
    Mug & Z & $\infty$ \\
    Pitcher Base & Z & $\infty$ \\
    Banana & - & - \\
    Power Drill & - & - \\
    \end{tabular}
    }
    % \vspace{-0.2cm}
    \caption{$\twohodataset$ objects and their axis and angle of symmetry considered during training and evaluation with our Keypoint Transformer.}
    \label{tab:sym}
    % \vspace{-0.3cm}
\end{table}

\begin{table}[b]
    \centering
    \scalebox{1.0}{
    \begin{tabular}{P{2.5cm}|P{2.5cm}}
    Object & MSSD (cm)  \\
    \hline
    Bleach Cleanser & 7.7 \\
    Mug & 6.5 \\
    Banana & 9.8 \\
    Pitcher Base & 7.9 \\
    Bowl & 7.8 \\
    Scissors & 13.5 \\
    Power Drill & 8.5 \\
    \hline
    All & 7.9
    \end{tabular}
    }
    % \vspace{-0.2cm}
    \caption{Object pose estimation accuracy of our Keypoint Transformer on the $\twohodataset$ dataset.}
    \label{tab:mssd_h2o3d}
    % \vspace{-0.3cm}
\end{table}
% DO NOT DELETE THIS
% Mean obj adi corner error 025_mug (count = 2092) = 0.073008 mts
% Mean obj adi corner error 011_banana (count = 154) = 0.109071 mts
% Mean obj adi corner error 019_pitcher_base (count = 900) = 0.072770 mts
% Mean obj adi corner error 024_bowl (count = 839) = 0.082044 mts
% Mean obj adi corner error 037_scissors (count = 22) = 0.065035 mts
% Mean obj adi corner error 035_power_drill (count = 343) = 0.072891 mts
% Mean obj corner adi all = 0.076069 mts

\section{Qualitative Results and Comparisons}
We provide here more qualitative results on HO-3D, $\twohodataset$ and InterHand2.6M.

\subsection{HO-3D and $\textbf{\twohodataset}$ Qualitative Results}
Fig.~\ref{fig:hand_objects_qualitative} shows qualitative results on $\twohodataset$ and HO-3D. Note that as we do not model contacts and interpenetration between hands and object, our method sometimes results in implausible poses as we show in the last example of Fig.~\ref{fig:hand_objects_qualitative}.

\subsection{InterHand2.6M Qualitative Results}
Fig.~\ref{fig:inter_comp} compares the estimated poses using the InterNet method from \cite{Moon_2020_ECCV_InterHand2.6M} and our proposed approach. As noted in Section~1 and Table~3 of the main paper, purely CNN-based  approaches do not explicitly model the relationship between image features of joints and tend to \textit{confuse}  joints during complex interactions. Our method performs well during complex interactions and strong occlusions (see last row of Fig.~\ref{fig:inter_comp}).

\begin{figure*}[t!]
\begin{center}
\includegraphics[trim=210 0 210 0,clip,width=1.0\linewidth]{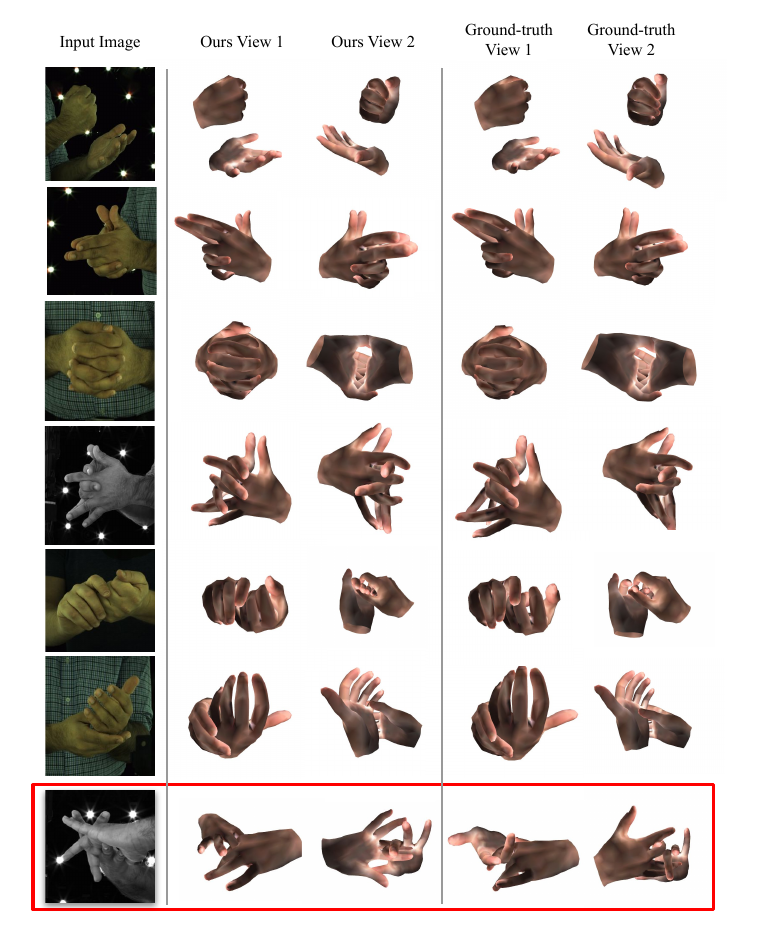}
\end{center}
% \vspace{-0.8cm}
\caption{Qualitative results of our method on InterHand2.6M~\cite{Moon_2020_ECCV_InterHand2.6M} compared to ground-truth poses. Our method predicts accurate poses in most scenarios. The last row shows a failure case where our method cannot recover the accurate pose due to complex pose and severe occlusion.}
\label{fig:inter_qualitative}
\vspace{-0.5cm}
\end{figure*}

We show more qualitative results using the MANO angle representation in Fig.~\ref{fig:inter_qualitative}. Our retrieved poses are very similar to ground-truth poses. As we show in the last row of Fig.~\ref{fig:inter_qualitative}, our method fails during scenarios where the hand is severely occluded during complex interaction.

\section{Attention Visualization}

\begin{figure*}
    \centering
    \includegraphics[trim=80 0 280 0,clip,width=1\linewidth]{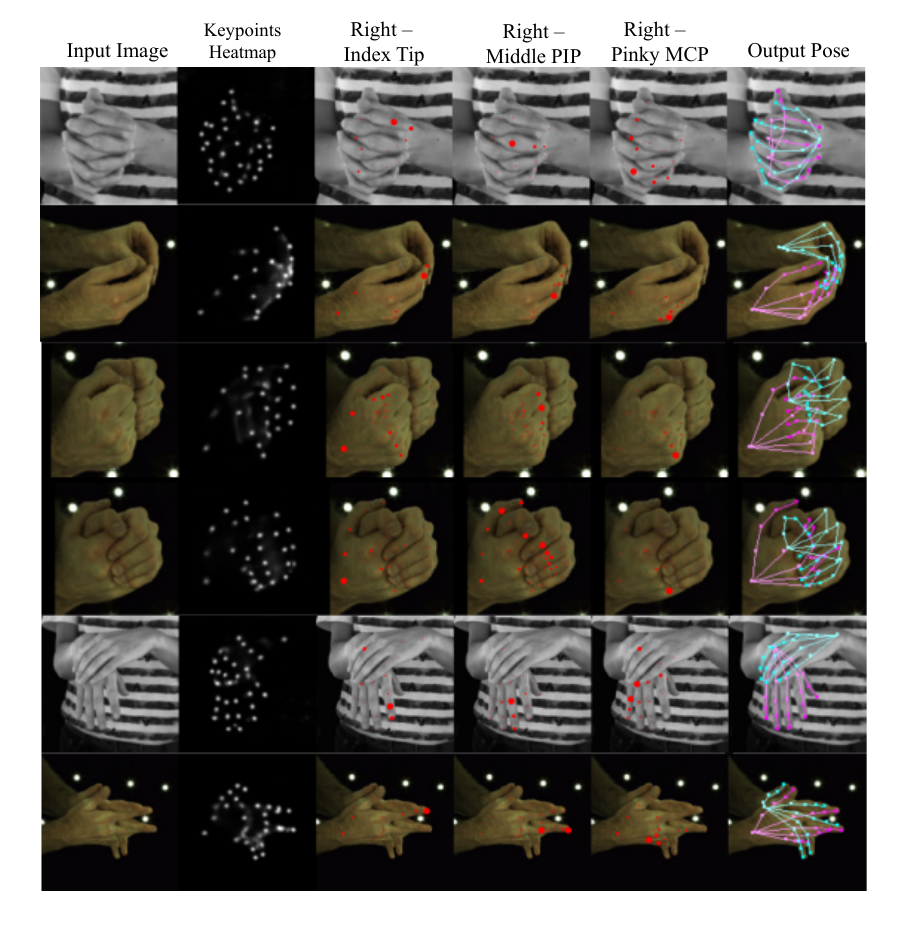}
    \caption{Attention visualization for 3 joint queries. Each joint query attends to the image feature from the respective joint location.}
    \label{fig:attn_vis}
\end{figure*}

In Fig.~\ref{fig:attn_vis}, we show more visualization of the cross-attention weights for three different joint queries. More specifically, the cross-attention weights represent the multiplicative factor on each of the keypoint features for a given joint query. We observe that the cross-attention learns to select keypoint(s) from respective joint location for each joint query when the joint is visible. For occluded joints, features from nearby visible joints are selected.

%%%%%%%%%%%%%%%%%%%%%%%%%%%%%%%%%%%%%%%%%%%%%%%%%%%%%%%%%
%%%%%%%%%%%%%%%%%%%%%%%%%%%%%%%%%%%%%%%%%%%%%%%%%%%%%%%%%

%%%%%%%%%%%%%%%%%%%%%%%%%%%%%%%%%%%%%%%%%%%%%%%%%%%%%%%%%
\newcommand{\nicerresultwidth}{0.13\linewidth}
\newcommand{\nicerrresult}[2]{
\includegraphics[width=\nicerresultwidth]{figures/supplementary/h2o_qual/#1_input.png} &
\includegraphics[width=\nicerresultwidth]{figures/supplementary/h2o_qual/#1_view1.png} &
\includegraphics[width=\nicerresultwidth]{figures/supplementary/h2o_qual/#1_view2.png} &
\includegraphics[width=\nicerresultwidth]{figures/supplementary/h2o_qual/#2_input.png} &
\includegraphics[width=\nicerresultwidth]{figures/supplementary/h2o_qual/#2_view1.png} &
\includegraphics[width=\nicerresultwidth]{figures/supplementary/h2o_qual/#2_view2.png} \\
}
\fboxsep=0mm%padding thickness
\fboxrule=1pt%border thickness
\begin{figure*}
    \centering
    \begin{tabular}{ccc|ccc}
        % RGB-D scan & Annotations from \cite{Avetisyan_2019_CVPR} & MCSS output & \textit{VoteNet Baseline}\\

        % \niceresult{scene0030_00}
        % \nicerresult{scene0500_00} 
        \nicerrresult{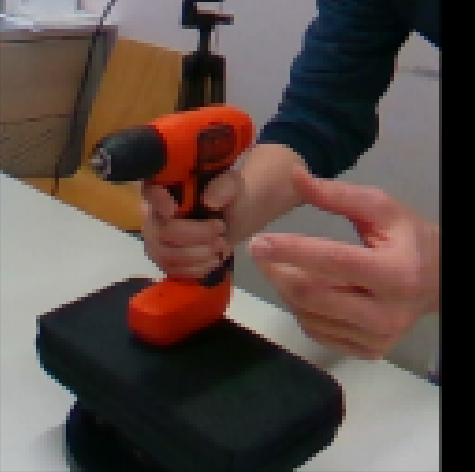}{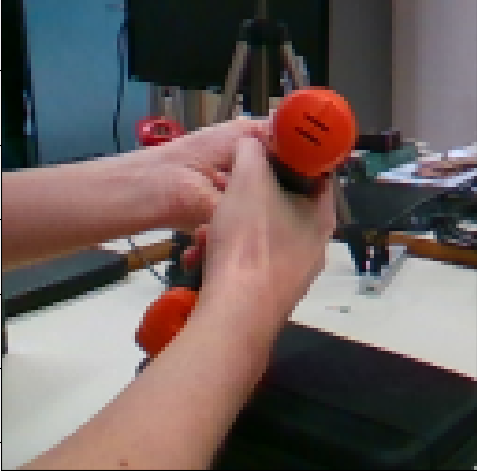}
        \nicerrresult{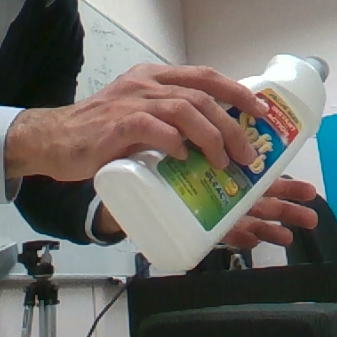}{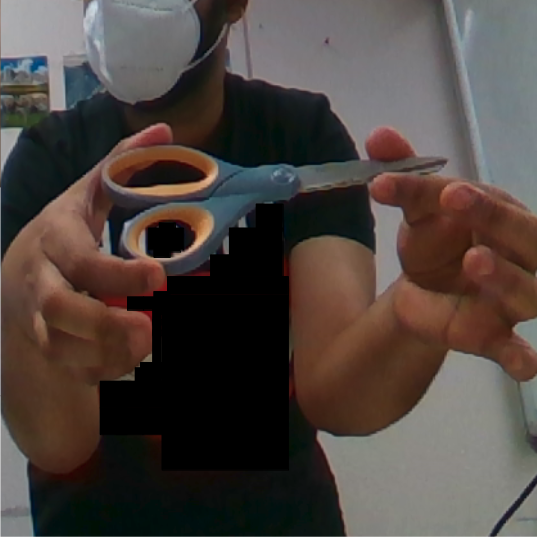}
        \nicerrresult{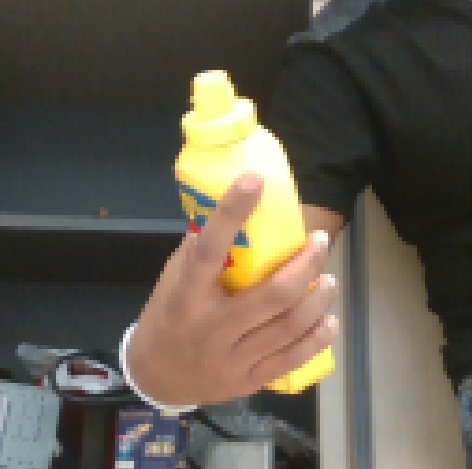}{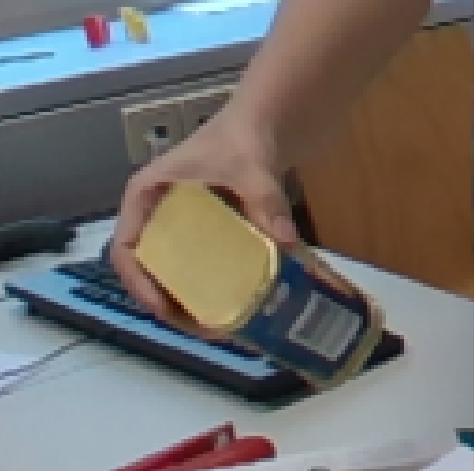}
        \includegraphics[width=\nicerresultwidth]{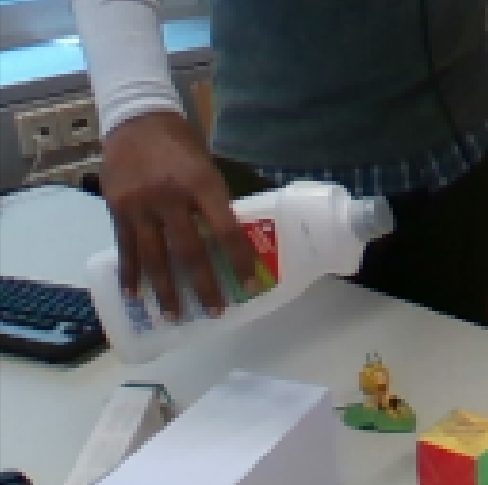} &
        \includegraphics[width=\nicerresultwidth]{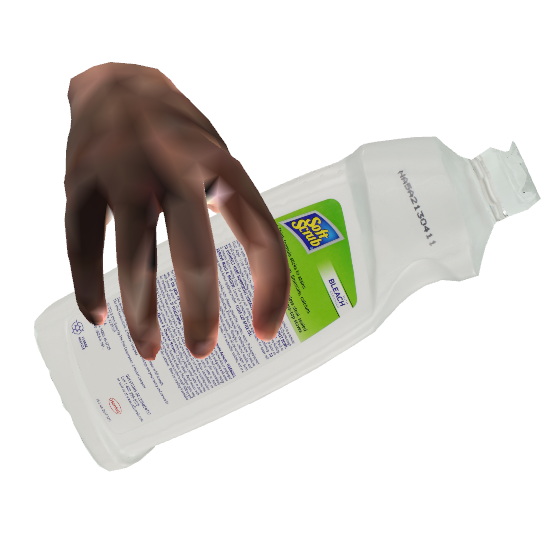} &
        \includegraphics[width=\nicerresultwidth]{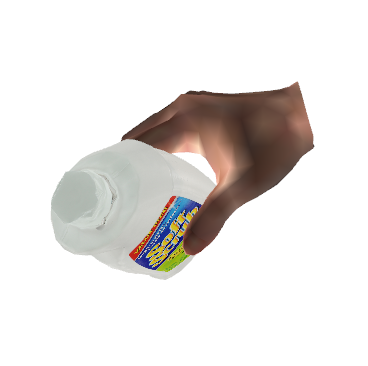} &
        \fcolorbox{red}{white}{
        \includegraphics[width=0.14\linewidth]{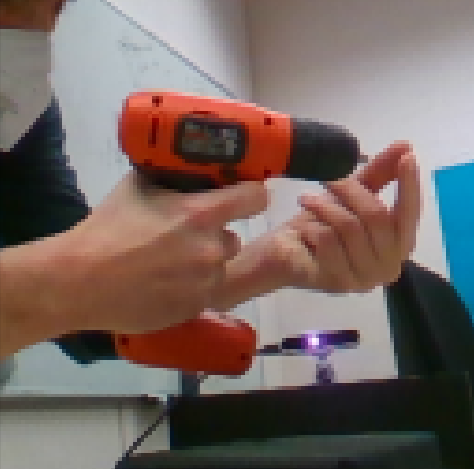}} &
        \fcolorbox{red}{white}{
        \includegraphics[width=0.14\linewidth]{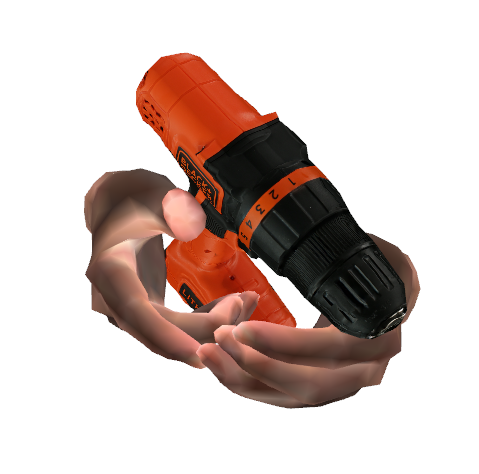}} &
        \fcolorbox{red}{white}{
        \includegraphics[width=0.14\linewidth]{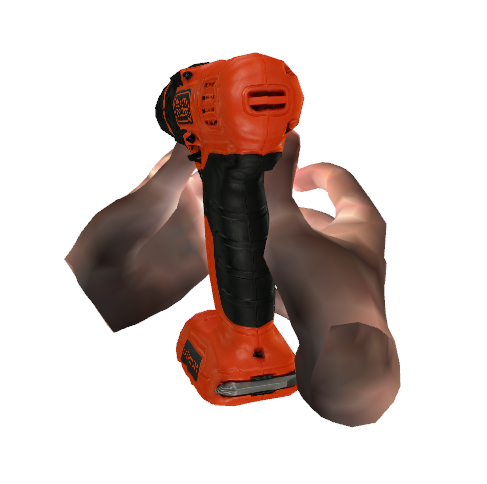}} \\

        % (a) & (b) & (c) & (d) \\
    \end{tabular}
    \caption{Qualitative results on $\twohodataset$ and HO-3D~\cite{hampali2020honnotate}. Our method obtains state-of-the-art results on HO-3D  while predicting reasonable results on $\twohodataset$. The last example is a failure case where the  predicted relative translations are inaccurate. }
    \label{fig:hand_objects_qualitative}
\end{figure*}
%%%%%%%%%%%%%%%%%%%%%%%%%%%%%%%%%%%%%%%%%%%%%%%%%%%%%%%%%

%%%%%%%%%%%%%%%%%%%%%%%%%%%%%%%%%%%%%%%%%%%%%%%%%%%%%%%%%

\newcommand{\intercompwidth}{0.19\linewidth}
\begin{figure*}
\centering
\begin{minipage}{\intercompwidth}
\centering
Input Image
\end{minipage}%
\begin{minipage}{\intercompwidth}
\centering
InterNet~\cite{Moon_2020_ECCV_InterHand2.6M} 2D Pose
\end{minipage}%
\begin{minipage}{\intercompwidth}
\centering
InterNet~\cite{Moon_2020_ECCV_InterHand2.6M} 3D Pose
\end{minipage}%
\begin{minipage}{\intercompwidth}
\centering
Ours 2D Pose
\end{minipage}%
\begin{minipage}{\intercompwidth}
\centering
Ours 3D Pose
\end{minipage}%

\par\medskip
\vspace{-5px}

\begin{minipage}{\intercompwidth}
\centering
\subfloat{\includegraphics[width=1\linewidth]{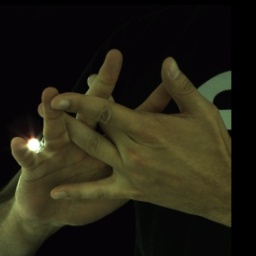}}
\end{minipage}%
\begin{minipage}{\intercompwidth}
\centering
\subfloat{\includegraphics[width=1\linewidth]{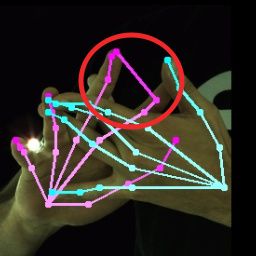}}
\end{minipage}%
\begin{minipage}{\intercompwidth}
\centering
\subfloat{\includegraphics[trim=90 50 90 60,clip,width=1\linewidth]{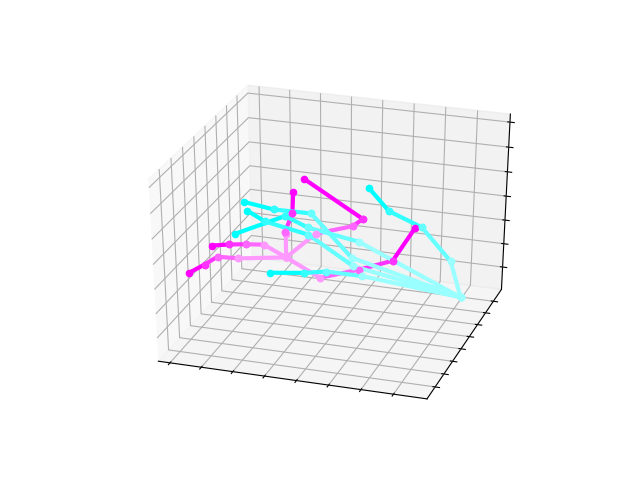}}
\end{minipage}%
\begin{minipage}{\intercompwidth}
\centering
\subfloat{\includegraphics[width=1\linewidth]{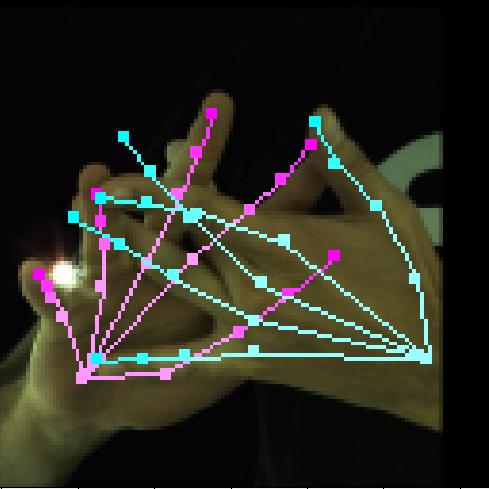}}
\end{minipage}%
\begin{minipage}{\intercompwidth}
\centering
\subfloat{\includegraphics[trim=90 50 90 60,clip,width=1\linewidth]{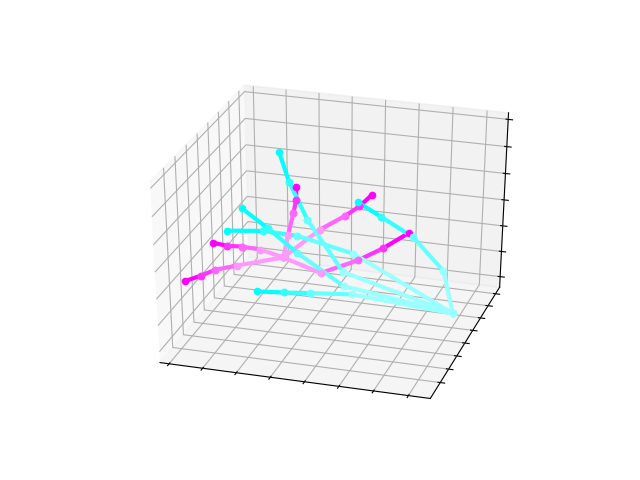}}
\end{minipage}

\par\medskip

\begin{minipage}{\intercompwidth}
\centering
\subfloat{\includegraphics[width=1\linewidth]{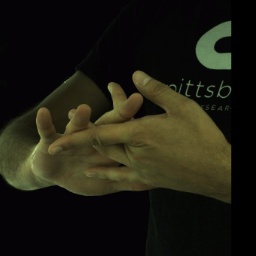}}
\end{minipage}%
\begin{minipage}{\intercompwidth}
\centering
\subfloat{\includegraphics[width=1\linewidth]{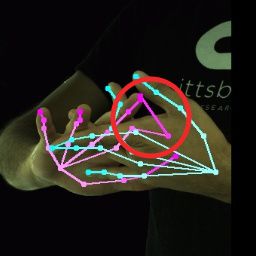}}
\end{minipage}%
\begin{minipage}{\intercompwidth}
\centering
\subfloat{\includegraphics[trim=90 50 90 60,clip,width=1\linewidth]{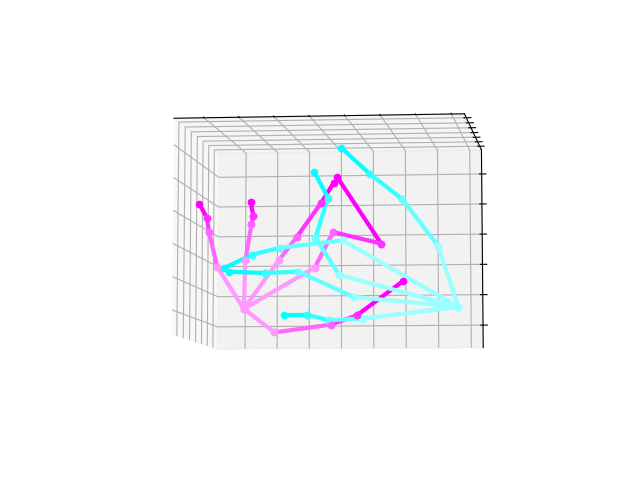}}
\end{minipage}%
\begin{minipage}{\intercompwidth}
\centering
\subfloat{\includegraphics[width=1\linewidth]{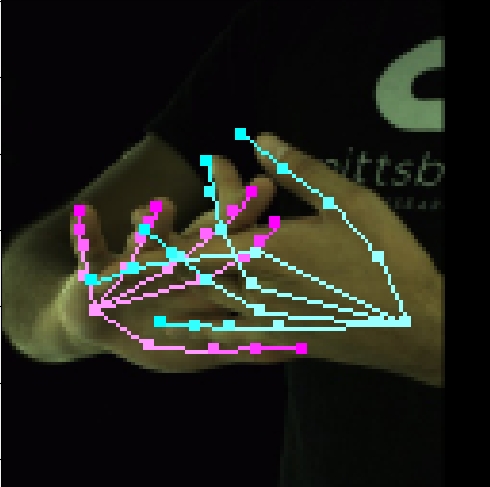}}
\end{minipage}%
\begin{minipage}{\intercompwidth}
\centering
\subfloat{\includegraphics[trim=90 50 90 60,clip,width=1\linewidth]{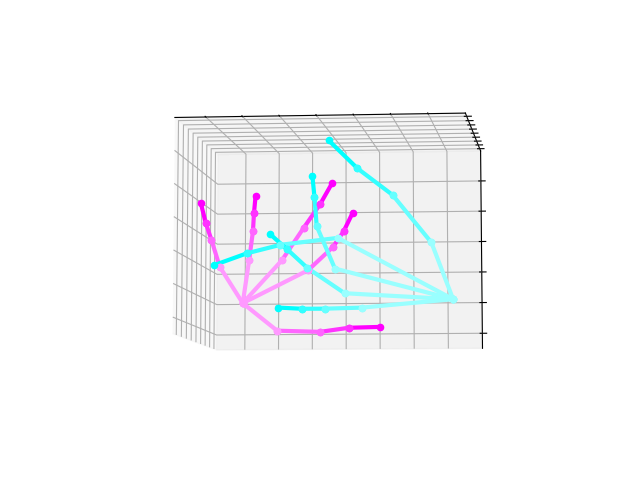}}
\end{minipage}

\par\medskip

\begin{minipage}{\intercompwidth}
\centering
\subfloat{\includegraphics[width=1\linewidth]{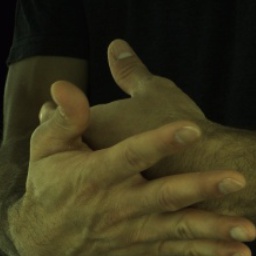}}
\end{minipage}%
\begin{minipage}{\intercompwidth}
\centering
\subfloat{\includegraphics[width=1\linewidth]{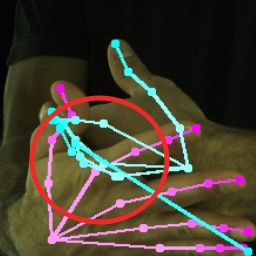}}
\end{minipage}%
\begin{minipage}{\intercompwidth}
\centering
\subfloat{\includegraphics[trim=90 50 90 60,clip,width=1\linewidth]{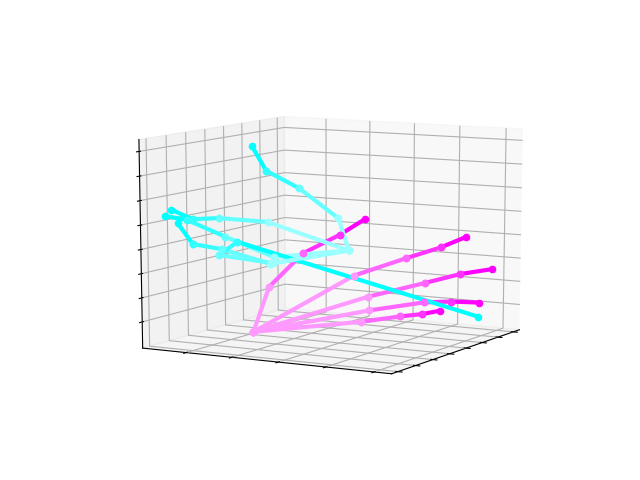}}
\end{minipage}%
\begin{minipage}{\intercompwidth}
\centering
\subfloat{\includegraphics[width=1\linewidth]{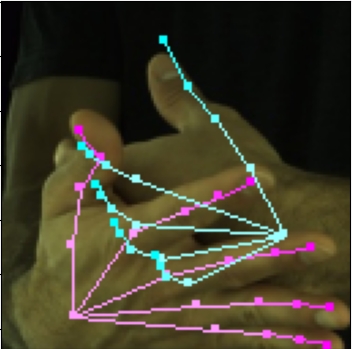}}
\end{minipage}%
\begin{minipage}{\intercompwidth}
\centering
\subfloat{\includegraphics[trim=90 50 90 60,clip,width=1\linewidth]{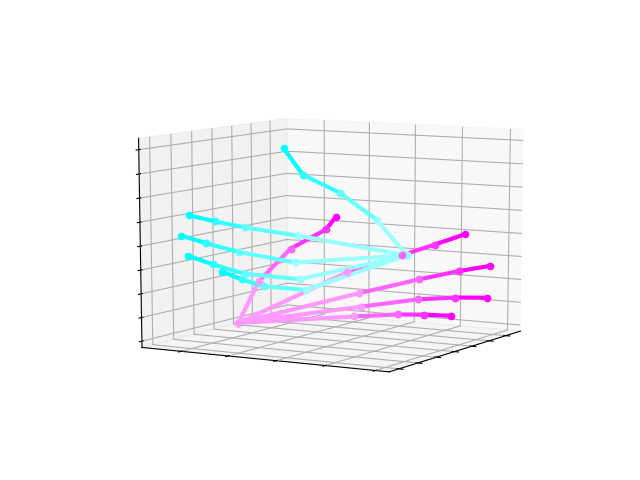}}
\end{minipage}

\par\medskip

\begin{minipage}{\intercompwidth}
\centering
\subfloat{\includegraphics[width=1\linewidth]{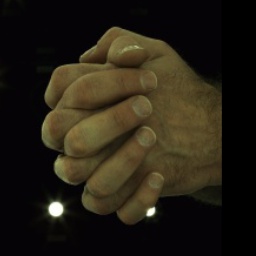}}
\end{minipage}%
\begin{minipage}{\intercompwidth}
\centering
\subfloat{\includegraphics[width=1\linewidth]{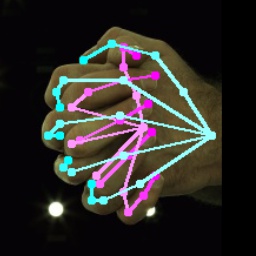}}
\end{minipage}%
\begin{minipage}{\intercompwidth}
\centering
\subfloat{\includegraphics[trim=90 50 90 60,clip,width=1\linewidth]{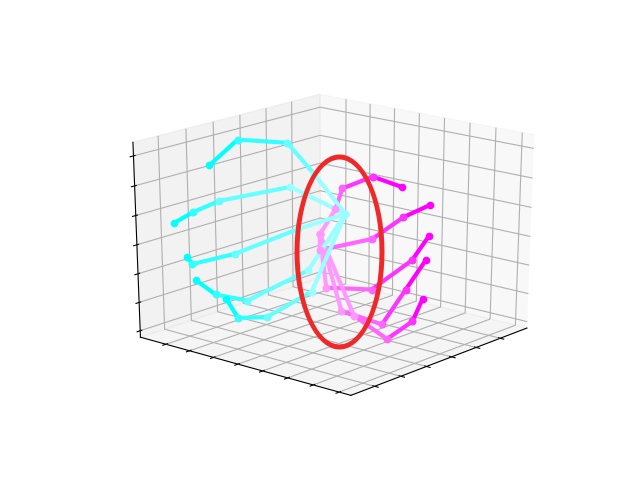}}
\end{minipage}%
\begin{minipage}{\intercompwidth}
\centering
\subfloat{\includegraphics[width=1\linewidth]{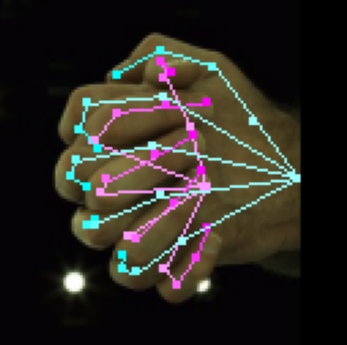}}
\end{minipage}%
\begin{minipage}{\intercompwidth}
\centering
\subfloat{\includegraphics[trim=90 50 90 60,clip,width=1\linewidth]{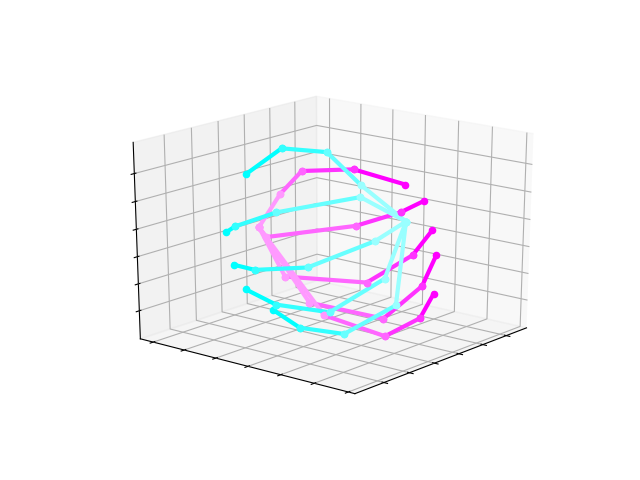}}
\end{minipage}

\par\medskip

\begin{minipage}{\intercompwidth}
\centering
\subfloat{\includegraphics[width=1\linewidth]{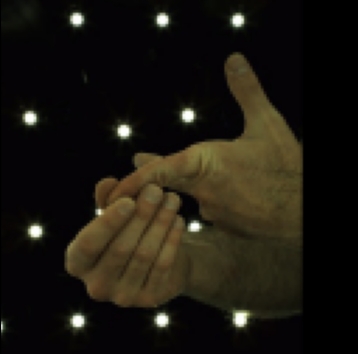}}
\end{minipage}%
\begin{minipage}{\intercompwidth}
\centering
\subfloat{\includegraphics[width=1\linewidth]{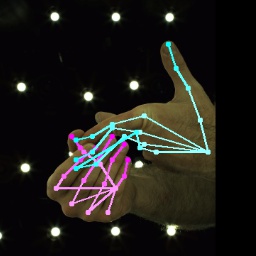}}
\end{minipage}%
\begin{minipage}{\intercompwidth}
\centering
\subfloat{\includegraphics[trim=90 50 90 60,clip,width=1\linewidth]{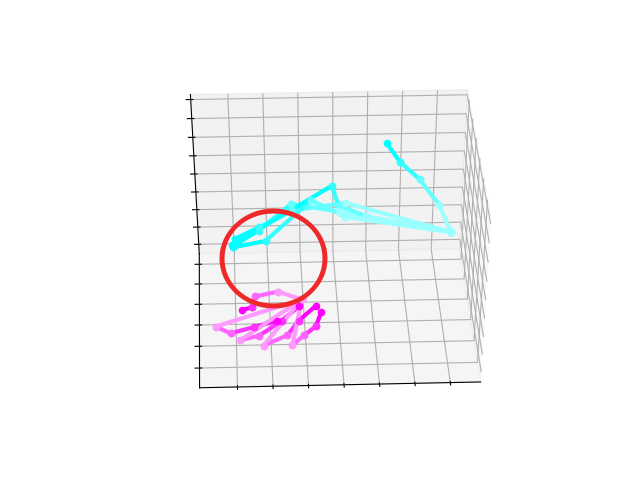}}
\end{minipage}%
\begin{minipage}{\intercompwidth}
\centering
\subfloat{\includegraphics[width=1\linewidth]{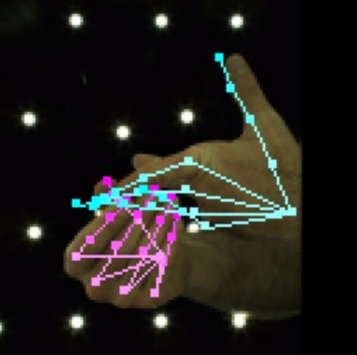}}
\end{minipage}%
\begin{minipage}{\intercompwidth}
\centering
\subfloat{\includegraphics[trim=90 50 90 60,clip,width=1\linewidth]{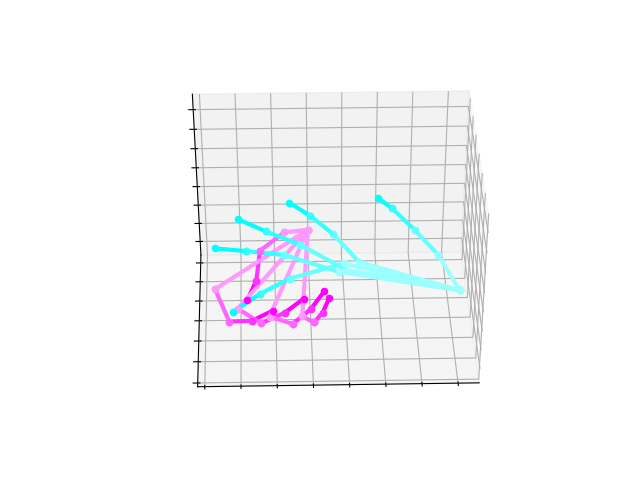}}
\end{minipage}

% \par\medskip

% \begin{minipage}{\intercompwidth}
% \centering
% \subfloat{\includegraphics[width=1\linewidth]{figures/supplementary/inter_compare/6_input.jpg}}
% \end{minipage}%
% \begin{minipage}{\intercompwidth}
% \centering
% \subfloat{\includegraphics[width=1\linewidth]{figures/supplementary/inter_compare/6_2d_inter.jpg}}
% \end{minipage}%
% \begin{minipage}{\intercompwidth}
% \centering
% \subfloat{\includegraphics[trim=90 50 90 60,clip,width=1\linewidth]{figures/supplementary/inter_compare/6_3d_inter.png}}
% \end{minipage}%
% \begin{minipage}{\intercompwidth}
% \centering
% \subfloat{\includegraphics[width=1\linewidth]{figures/supplementary/inter_compare/6_2d_ours.jpg}}
% \end{minipage}%
% \begin{minipage}{\intercompwidth}
% \centering
% \subfloat{\includegraphics[trim=90 50 90 60,clip,width=1\linewidth]{figures/supplementary/inter_compare/6_3d_ours.png}}
% \end{minipage}

\caption{Qualitative comparison between InterNet~\cite{Moon_2020_ECCV_InterHand2.6M} and our proposed method. Our method outputs more accurate poses even during strong occlusions. Red circles indicate regions where InterNet results are inaccurate.}
\label{fig:inter_comp}
\vspace{-0.5cm}
\end{figure*}

%%%%%%%%%%%%%%%%%%%%%%%%%%%%%%%%%%%%%%%%%%%%%%%%%%%%%%%%%
%%%%%%%%%%%%%%%%%%%%%%%%%%%%%%%%%%%%%%%%%%%%%%%%%%%%%%%%%

%%%%%%%%%%%%%%%%%%%%%%%%%%%%%%%%%%%%%%%%%%%%%%%%%%%%%%%%%

%%%%%%%%%%%%%%%%%%%%%%%%%%%%%%%%%%%%%%%%%%%%%%%%%%%%%%%%%

%%%%%%%%%%%%%%%%%%%%%%%%%%%%%%%%%%%%%%%%%%%%%%%%%%%

{\small
\bibliographystyle{ieee_fullname}
% \bibliography{egbib}
\bibliography{biblio}

\begin{thebibliography}{10}\itemsep=-1pt

\bibitem{Armagan2020MeasuringGT}
Anil Armagan, Guillermo Garcia-Hernando, Seungryul Baek, Shreyas Hampali, Mahdi
  Rad, Zhaohui Zhang, Shipeng Xie, Ming-xiu Chen, Boshen Zhang, F. Xiong, Yang
  Xiao, Zhiguo Cao, Junsong Yuan, Pengfei Ren, Weiting Huang, Haifeng Sun,
  Marek Hr{\'u}z, Jakub Kanis, Zdenek Krnoul, Qingfu Wan, Shile Li, Linlin
  Yang, Dongheui Lee, Angela Yao, Weiguo Zhou, Sijia Mei, Yunhui Liu, Adrian
  Spurr, Umar Iqbal, Pavlo Molchanov, Philippe Weinzaepfel, Romain Br{\'e}gier,
  Gr{\'e}gory Rogez, Vincent Lepetit, and Tae-Kyun Kim.
\newblock {Measuring Generalisation to Unseen Viewpoints, Articulations, Shapes
  and Objects for 3D Hand Pose Estimation Under Hand-Object Interaction}.
\newblock In {\em European Conference on Computer Vision}, 2020.

\bibitem{Baek2019PushingTE}
Seungryul Baek, Kwang~In Kim, and Tae-Kyun Kim.
\newblock {Pushing the Envelope for RGB-Based Dense 3D Hand Pose Estimation via
  Neural Rendering}.
\newblock In {\em Conference on Computer Vision and Pattern Recognition}, 2019.

\bibitem{Baek2020WeaklySupervisedDA}
Seungryul Baek, Kwang~In Kim, and Tae-Kyun Kim.
\newblock {Weakly-Supervised Domain Adaptation via GAN and Mesh Model for
  Estimating 3D Hand Poses Interacting Objects}.
\newblock In {\em Conference on Computer Vision and Pattern Recognition}, 2020.

\bibitem{Ballan2012MotionCO}
Luca Ballan, Aparna Taneja, Juergen Gall, Luc Van~Gool, and Marc Pollefeys.
\newblock {Motion Capture of Hands in Action Using Discriminative Salient
  Points}.
\newblock In {\em European Conference on Computer Vision}, 2012.

\bibitem{Brahmbhatt2020ContactPoseAD}
Samarth Brahmbhatt, Chengcheng Tang, Christopher~D. Twigg, Charles~C. Kemp, and
  James Hays.
\newblock {ContactPose: A Dataset of Grasps with Object Contact and Hand Pose}.
\newblock In {\em European Conference on Computer Vision}, 2020.

\bibitem{guillem2021}
Guillem Braso, Nikita Kister, and Laura Leal-Taix\'e.
\newblock {The Center of Attention: Center-Keypoint Grouping Attention for
  Multi-Person Pose Estimation}.
\newblock In {\em International Conference on Computer Vision}, 2021.

\bibitem{Cao2020ReconstructingHI}
Zhe Cao, Ilija Radosavovic, Angjoo Kanazawa, and Jitendra Malik.
\newblock {Reconstructing Hand-Object Interactions in the Wild}.
\newblock In {\em arXiv Preprint}, 2020.

\bibitem{detr}
Nicolas Carion, Francisco Massa, Gabriel Synnaeve, Nicolas Usunier, Alexander
  Kirillov, and Sergey Zagoruyko.
\newblock {End-to-End Object Detection with Transformers}.
\newblock In {\em European Conference on Computer Vision}, 2020.

\bibitem{i2v}
Ping Chen, Yujin Chen, Dong Yang, Fangyin Wu, Qin Li, Qingpei Xia, and Yong
  Tan.
\newblock I2uv-handnet: Image-to-uv prediction network for accurate and
  high-fidelity 3d hand mesh modeling.
\newblock In {\em International Conference on Computer Vision}, 2021.

\bibitem{pose2mesh}
Hongsuk Choi, Gyeongsik Moon, and Kyoung~Mu Lee.
\newblock Pose2mesh: Graph convolutional network for 3d human pose and mesh
  recovery from a 2d human pose.
\newblock In {\em European Conference on Computer Vision (ECCV)}, 2020.

\bibitem{Dosovitskiy2020AnII}
Alexey Dosovitskiy, Lucas Beyer, Alexander Kolesnikov, Dirk Weissenborn,
  Xiaohua Zhai, Thomas Unterthiner, Mostafa Dehghani, Matthias Minderer, Georg
  Heigold, Sylvain Gelly, Jakob Uszkoreit, and Neil Houlsby.
\newblock {An Image Is Worth 16x16 Words: Transformers for Image Recognition at
  Scale}.
\newblock In {\em arXiv Preprint}, 2020.

\bibitem{fan2021digit}
Zicong Fan, Adrian Spurr, Muhammed Kocabas, Siyu Tang, Michael Black, and Otmar
  Hilliges.
\newblock {Learning to Disambiguate Strongly Interacting Hands via
  Probabilistic Per-Pixel Part Segmentation}.
\newblock In {\em International Conference on 3D Vision}, 2021.

\bibitem{GarciaHernando2018FirstPersonHA}
Guillermo Garcia-Hernando, Shanxin Yuan, Seungryul Baek, and Tae-Kyun Kim.
\newblock {First-Person Hand Action Benchmark with RGB-D Videos and 3D Hand
  Pose Annotations}.
\newblock In {\em Conference on Computer Vision and Pattern Recognition}, 2018.

\bibitem{hampali2020honnotate}
Shreyas Hampali, Mahdi Rad, Markus Oberweger, and Vincent Lepetit.
\newblock {HOnnotate: A Method for 3D Annotation of Hand and Object Poses}.
\newblock In {\em Conference on Computer Vision and Pattern Recognition}, 2020.

\bibitem{Han2020MEgATrackME}
Shangchen Han, Beibei Liu, R. Cabezas, Christopher~D. Twigg, Peizhao Zhang,
  Jeff Petkau, Tsz-Ho Yu, Chun-Jung Tai, Muzaffer Akbay, Zheng Wang, Asaf
  Nitzan, Gang Dong, Yuting Ye, Lingling Tao, Chengde Wan, and Robert Wang.
\newblock {MEgATrack: Monochrome Egocentric Articulated Hand-Tracking for
  Virtual Reality}.
\newblock {\em IEEE Transactions on Robotics and Automation}, 39, 2020.

\bibitem{hasson20_handobjectconsist}
Yana Hasson, Bugra Tekin, Federica Bogo, Ivan Laptev, Marc Pollefeys, and
  Cordelia Schmid.
\newblock {Leveraging Photometric Consistency over Time for Sparsely Supervised
  Hand-Object Reconstruction}.
\newblock In {\em Conference on Computer Vision and Pattern Recognition}, 2020.

\bibitem{hasson19_obman}
Yana Hasson, G{\"u}l Varol, Dimitrios Tzionas, Igor Kalevatykh, Michael~J.
  Black, Ivan Laptev, and Cordelia Schmid.
\newblock {Learning Joint Reconstruction of Hands and Manipulated Objects}.
\newblock In {\em Conference on Computer Vision and Pattern Recognition}, 2019.

\bibitem{He2016DeepRL}
Kaiming He, X. Zhang, Shaoqing Ren, and Jian Sun.
\newblock {Deep Residual Learning for Image Recognition}.
\newblock In {\em Conference on Computer Vision and Pattern Recognition}, 2016.

\bibitem{HodanSDLBMRM20}
Tom{\'{a}}s Hodan, Martin Sundermeyer, Bertram Drost, Yann Labb{\'{e}}, Eric
  Brachmann, Frank Michel, Carsten Rother, and Jiri Matas.
\newblock {{BOP} Challenge 2020 on 6D Object Localization}.
\newblock In {\em Computer Vision - {ECCV} 2020 Workshops - Glasgow, UK, August
  23-28, 2020, Proceedings, Part {II}}, 2020.

\bibitem{Huang2020HandTransformerNS}
Lin Huang, Jianchao Tan, Ji Liu, and Junsong Yuan.
\newblock {Hand-Transformer: Non-Autoregressive Structured Modeling for 3D Hand
  Pose Estimation}.
\newblock In {\em European Conference on Computer Vision}, 2020.

\bibitem{umar}
Umar Iqbal, Pavlo Molchanov, Thomas Breuel, Juergen Gall, and Jan Kautz.
\newblock {Hand Pose Estimation via Latent 2.5D Heatmap Regression}.
\newblock In {\em European Conference on Computer Vision}, 2018.

\bibitem{Angjoo}
Angjoo Kanazawa, Michael~J. Black, David~W. Jacobs, and Jitendra Malik.
\newblock {End-to-End Recovery of Human Shape and Pose}.
\newblock In {\em Conference on Computer Vision and Pattern Recognition}, 2018.

\bibitem{Karunratanakul2020GraspingFL}
Korrawe Karunratanakul, Jinlong Yang, Yan Zhang, Michael~J. Black, Krikamol
  Muandet, and Siyu Tang.
\newblock {Grasping Field: Learning Implicit Representations for Human Grasps}.
\newblock In {\em International Conference on 3D Vision}, 2020.

\bibitem{Khan2021TransformersIV}
Salman Khan, Muzammal Naseer, Munawar Hayat, Syed~Waqas Zamir, Fahad Khan, and
  Mubarak Shah.
\newblock {Transformers in Vision: A Survey}.
\newblock In {\em arXiv Preprint}, 2021.

\bibitem{Dong}
Dong~Uk Kim, Kwang~In Kim, and Seungryul Baek.
\newblock {End-to-End Detection and Pose Estimation of Two Interacting Hands}.
\newblock In {\em International Conference on Computer Vision}, 2021.

\bibitem{KingmaB14}
Diederik~P. Kingma and Jimmy Ba.
\newblock {Adam: {A} Method for Stochastic Optimization}.
\newblock In {\em International Conference for Learning Representations}, 2015.

\bibitem{Kulon2020WeaklySupervisedMH}
Dominik Kulon, Riza~Alp G{\"u}ler, Iasonnas Kokkinos, Michael Bronstein, and
  Stefanos Zafeiriou.
\newblock {Weakly-Supervised Mesh-Convolutional Hand Reconstruction in the
  Wild}.
\newblock In {\em Conference on Computer Vision and Pattern Recognition}, 2020.

\bibitem{Kumar2021ColorizationT}
Manoj Kumar, Dirk Weissenborn, and Nal Kalchbrenner.
\newblock {Colorization Transformer}.
\newblock In {\em arXiv Preprint}, 2021.

\bibitem{Kwon_2021_ICCV}
Taein Kwon, Bugra Tekin, Jan St\"uhmer, Federica Bogo, and Marc Pollefeys.
\newblock {H2O: Two Hands Manipulating Objects for First Person Interaction
  Recognition}.
\newblock In {\em International Conference on Computer Vision}, 2021.

\bibitem{Kyriazis2014Scalable3T}
Nikolaos Kyriazis and Antonis~A. Argyros.
\newblock {Scalable 3D Tracking of Multiple Interacting Objects}.
\newblock In {\em Conference on Computer Vision and Pattern Recognition}, 2014.

\bibitem{artiboost}
Kailin Li, Lixin Yang, Xinyu Zhan, Jun Lv, Wenqiang Xu, Jiefeng Li, and Cewu
  Lu.
\newblock {ArtiBoost}: Boosting articulated 3d hand-object pose estimation via
  online exploration and synthesis.
\newblock In {\em Conference on Computer Vision and Pattern Recognition}, 2022.

\bibitem{metro}
Kevin Lin, Lijuan Wang, and Zicheng Liu.
\newblock {End-to-End Human Pose and Mesh Reconstruction with Transformers}.
\newblock In {\em Conference on Computer Vision and Pattern Recognition}, 2021.

\bibitem{lin2021}
Kevin Lin, Lijuan Wang, and Zicheng Liu.
\newblock {Mesh Graphormer}.
\newblock In {\em International Conference on Computer Vision}, 2021.

\bibitem{liu}
Shaowei Liu, Hanwen Jiang, Jiarui Xu, Sifei Liu, and Xiaolong Wang.
\newblock {Semi-Supervised 3D Hand-Object Poses Estimation with Interactions in
  Time}.
\newblock In {\em Conference on Computer Vision and Pattern Recognition}, 2021.

\bibitem{Moon2019CameraDT}
Gyeongsik Moon, Ju~Yong Chang, and Kyoung~Mu Lee.
\newblock {Camera Distance-Aware Top-Down Approach for 3D Multi-Person Pose
  Estimation from a Single RGB Image}.
\newblock In {\em International Conference on Computer Vision}, 2019.

\bibitem{Moon2020I2LMeshNetIP}
Gyeongsik Moon and Kyoung~Mu Lee.
\newblock {I2L-MeshNet: Image-to-Lixel Prediction Network for Accurate 3D Human
  Pose and Mesh Estimation from a Single RGB Image}.
\newblock In {\em European Conference on Computer Vision}, 2020.

\bibitem{Moon_2020_ECCV_InterHand2.6M}
Gyeongsik Moon, Shoou-I Yu, He Wen, Takaaki Shiratori, and Kyoung~Mu Lee.
\newblock {InterHand2.6M: A Dataset and Baseline for 3D Interacting Hand Pose
  Estimation from a Single RGB Image}.
\newblock In {\em European Conference on Computer Vision}, 2020.

\bibitem{Mueller2019RealtimePA}
Franziska Mueller, Micah Davis, Florian Bernard, Oleksandr Sotnychenko, Mickeal
  Verschoor, Miguel Otaduy, Dan Casas, and Christian Theobalt.
\newblock {Real-Time Pose and Shape Reconstruction of Two Interacting Hands
  with a Single Depth Camera}.
\newblock {\em IEEE Transactions on Robotics and Automation}, 38, 2019.

\bibitem{Oberweger2015TrainingAF}
Markus Oberweger, Paul Wohlhart, and Vincent Lepetit.
\newblock {Training a Feedback Loop for Hand Pose Estimation}.
\newblock In {\em International Conference on Computer Vision}, 2015.

\bibitem{Oikonomidis2011Tracking}
Iason Oikonomidis, Nikolaos Kyriazis, and Antonis~A. Argyros.
\newblock {Full DOF Tracking of a Hand Interacting with an Object by Modeling
  Occlusions and Physical Constraints}.
\newblock In {\em International Conference on Computer Vision}, 2011.

\bibitem{Oikonomidis2012TrackingTA}
I. Oikonomidis, Nikolaos Kyriazis, and Antonis~A. Argyros.
\newblock {Tracking the Articulated Motion of Two Strongly Interacting Hands}.
\newblock In {\em Conference on Computer Vision and Pattern Recognition}, 2012.

\bibitem{Panteleris2015}
Paschalis Panteleris, Nikolaos Kyriazis, and Antonis~A. Argyros.
\newblock {3D Tracking of Human Hands in Interaction with Unknown Objects}.
\newblock In {\em British Machine Vision Conference}, 2015.

\bibitem{Panteleris2018UsingAS}
Paschalis Panteleris, Iason Oikonomidis, and Antonis~A. Argyros.
\newblock {Using a Single RGB Frame for Real Time 3D Hand Pose Estimation in
  the Wild}.
\newblock In {\em IEEE Winter Conference on Applications of Computer Vision},
  2018.

\bibitem{handoccnet}
JoonKyu Park, Yeonguk Oh, Gyeongsik Moon, Hongsuk Choi, and Kyoung~Mu Lee.
\newblock {HandOccNet: Occlusion-Robust 3D Hand Mesh Estimation Network}.
\newblock In {\em Conference on Computer Vision and Pattern Recognition}, 2022.

\bibitem{Park2019Pix2PosePC}
Kiru Park, Timothy Patten, and Markus Vincze.
\newblock {Pix2Pose: Pixel-Wise Coordinate Regression of Objects for 6D Pose
  Estimation}.
\newblock In {\em International Conference on Computer Vision}, 2019.

\bibitem{Pavlakos2019TexturePoseSH}
Georgios Pavlakos, Nikos Kolotouros, and Kostas Daniilidis.
\newblock {TexturePose: Supervising Human Mesh Estimation with Texture
  Consistency}.
\newblock In {\em International Conference on Computer Vision}, 2019.

\bibitem{pavlakos17object3d}
Georgios Pavlakos, Xiaowei Zhou, Aaron Chan, Konstantinos~G Derpanis, and
  Kostas Daniilidis.
\newblock 6-dof object pose from semantic keypoints.
\newblock In {\em International Conference on Robotics and Automation (ICRA)},
  2017.

\bibitem{Pavlakos2018LearningTE}
Georgios Pavlakos, Luyang Zhu, Xiaowei Zhou, and Kostas Daniilidis.
\newblock {Learning to Estimate 3D Human Pose and Shape from a Single Color
  Image}.
\newblock In {\em Conference on Computer Vision and Pattern Recognition}, 2018.

\bibitem{Romero2017EmbodiedH}
Javier Romero, Dimitrios Tzionas, and Michael~J. Black.
\newblock {EMbodied Hands: Modeling and Capturing Hands and Bodies Together}.
\newblock {\em IEEE Transactions on Robotics and Automation}, 36, 2017.

\bibitem{unet}
Olaf Ronneberger, Philipp Fischer, and Thomas Brox.
\newblock {U-Net: Convolutional Networks for Biomedical Image Segmentation}.
\newblock In {\em Conference on Medical Image Computing and Computer Assisted
  Intervention}, 2015.

\bibitem{Smith2020ConstrainingDH}
Breannan Smith, Chenglei Wu, He Wen, Patrick Peluse, Yaser Sheikh, Jessica
  Hodgins, and Takaaki Shiratori.
\newblock {Constraining Dense Hand Surface Tracking with Elasticity}.
\newblock {\em IEEE Transactions on Robotics and Automation}, 39, 2020.

\bibitem{Adrian}
Adrian Spurr, Umar Iqbal, Pavlo Molchanov, Otmar Hilliges, and Jan Kautz.
\newblock {Weakly Supervised 3D Hand Pose Estimation via Biomechanical
  Constraints}.
\newblock In {\em European Conference on Computer Vision}, 2020.

\bibitem{Sridhar2016RealTimeJT}
Srinath Sridhar, Franziska Mueller, Michael Zollh{\"o}fer, Dan Casas, Antti
  Oulasvirta, and Christian Theobalt.
\newblock {Real-Time Joint Tracking of a Hand Manipulating an Object from RGB-D
  Input}.
\newblock In {\em European Conference on Computer Vision}, 2016.

\bibitem{Taheri2020GRABAD}
Omid Taheri, Nima Ghorbani, Michael~J. Black, and Dimitrios Tzionas.
\newblock {GRAB: A Dataset of Whole-Body Human Grasping of Objects}.
\newblock In {\em European Conference on Computer Vision}, 2020.

\bibitem{Taylor2016EfficientAP}
Jonathan Taylor, Lucas Bordeaux, Thomas Cashman, Bob Corish, Cem Keskin, Toby
  Sharp, Eduardo Soto, David Sweeney, Julien Valentin, Benjamin Luff, Arran
  Topalian, Erroll Wood, Sameh Khamis, Pushmeet Kohli, Shahram Izadi, Richard
  Banks, Andrew Fitzgibbon, and Jamie Shotton.
\newblock {Efficient and Precise Interactive Hand Tracking through Joint,
  Continuous Optimization of Pose and Correspondences}.
\newblock {\em IEEE Transactions on Robotics and Automation}, 35, 2016.

\bibitem{Taylor2017ArticulatedDF}
Jonathan Taylor, Vladimir Tankovich, Danhang Tang, Cem Keskin, David Kim,
  Philip Davidson, Adarsh Kowdle, and Shahram Izadi.
\newblock {Articulated Distance Fields for Ultra-Fast Tracking of Hands
  Interacting}.
\newblock {\em IEEE Transactions on Robotics and Automation}, 36, 2017.

\bibitem{Tekin2019HOUE}
Bugra Tekin, Federica Bogo, and Marc Pollefeys.
\newblock {H+O: Unified Egocentric Recognition of 3D Hand-Object Poses and
  Interactions}.
\newblock In {\em Conference on Computer Vision and Pattern Recognition}, 2019.

\bibitem{Tung2017SelfsupervisedLO}
Hsiao-Yu Tung, Hsiao-Wei Tung, Ersin Yumer, and Katerina Fragkiadaki.
\newblock {Self-Supervised Learning of Motion Capture}.
\newblock In {\em Advances in Neural Information Processing Systems}, 2017.

\bibitem{Tzionas2016CapturingHI}
Dimitrios Tzionas, Luca Ballan, Abhilash Srikantha, Pablo Aponte, Marc
  Pollefeys, and Juergen Gall.
\newblock {Capturing Hands in Action Using Discriminative Salient Points and
  Physics Simulation}.
\newblock {\em International Journal of Computer Vision}, 118, 2016.

\bibitem{Tzionas20153DOR}
Dimitrios Tzionas and Juergen Gall.
\newblock {3D Object Reconstruction from Hand-Object Interactions}.
\newblock In {\em International Conference on Computer Vision}, 2015.

\bibitem{Vaswani}
Ashish Vaswani, Noam Shazeer, Niki Parmar, Jakob Uszkoreit, Llion Jones,
  Aidan~N. Gomez, Lukasz Kaiser, and Illia Polosukhin.
\newblock {Attention Is All You Need}.
\newblock In {\em Advances in Neural Information Processing Systems}, 2017.

\bibitem{Wang2020RGB2HandsRT}
Jiayi Wang, Franziska Mueller, Florian Bernard, Suzanne Sorli, Oleksandr
  Sotnychenko, Neng Qian, Miguel Otaduy, Dan Casas, and Christian Theobalt.
\newblock {RGB2Hands: Real-Time Tracking of 3D Hand Interactions from Monocular
  RGB Video}.
\newblock {\em IEEE Transactions on Robotics and Automation}, 39, 2020.

\bibitem{Xiang2019MonocularTC}
Donglai Xiang, Hanbyul Joo, and Yaser Sheikh.
\newblock {Monocular Total Capture: Posing Face, Body, and Hands in the Wild}.
\newblock In {\em Conference on Computer Vision and Pattern Recognition}, 2019.

\bibitem{ycb}
Yu Xiang, Tanner Schmidt, Venkatraman Narayanan, and Dieter Fox.
\newblock {PoseCNN: A Convolutional Neural Network for 6D Object Pose
  Estimation in Cluttered Scenes}.
\newblock {\em Science}, 2018.

\bibitem{Yang2020LearningTT}
Fuzhi Yang, Huan Yang, Jianlong Fu, Hongtao Lu, and Baining Guo.
\newblock {Learning Texture Transformer Network for Image Super-Resolution}.
\newblock In {\em Conference on Computer Vision and Pattern Recognition}, 2020.

\bibitem{Baowen}
Baowen Zhang, Yangang Wang, Xiaoming Deng, Yinda Zhang, Ping Tan, Cuixia Ma,
  and Hongan Wang.
\newblock {Interacting Two-Hand 3D Pose and Shape Reconstruction from Single
  Color Image}.
\newblock In {\em International Conference on Computer Vision}, 2021.

\bibitem{zheng}
Xiaozheng Zheng, Pengfei Ren, Haifeng Sun, Jingyu Wang, Qi Qi, and Jianxin
  Liao.
\newblock Joint-aware regression: Rethinking regression-based method for 3d
  hand pose estimation.
\newblock In {\em British Machine Vision Conference}, 2021.

\bibitem{Zhou2019OnTC}
Y. Zhou, Connelly Barnes, Jingwan Lu, Jimei Yang, and Hao Li.
\newblock {On the Continuity of Rotation Representations in Neural Networks}.
\newblock In {\em Conference on Computer Vision and Pattern Recognition}, 2019.

\bibitem{zhu2021deformable}
Xizhou Zhu, Weijie Su, Lewei Lu, Bin Li, Xiaogang Wang, and Jifeng Dai.
\newblock {Deformable DETR: Deformable Transformers for End-to-End Object
  Detection}.
\newblock In {\em International Conference for Learning Representations}, 2021.

\bibitem{Zimmermann2017LearningTE}
Christian Zimmermann and Thomas Brox.
\newblock {Learning to Estimate 3D Hand Pose from Single RGB Images}.
\newblock In {\em International Conference on Computer Vision}, 2017.

\bibitem{Zimmermann2019FreiHANDAD}
Christian Zimmermann, Duygu Ceylan, Jimei Yang, Bryan~C. Russell, Max Argus,
  and Thomas Brox.
\newblock {FreiHAND: A Dataset for Markerless Capture of Hand Pose and Shape
  from Single RGB Images}.
\newblock In {\em International Conference on Computer Vision}, 2019.

\end{thebibliography}
}

\end{document}